\documentclass{article} 
\usepackage{iclr2024_conference,times}

\usepackage{tcolorbox}
\tcbuselibrary{breakable}
\usepackage{hyperref}
\usepackage{url}
\usepackage{color}
\usepackage{amsmath}    
\usepackage{tabularray}
\usepackage{multirow}
\usepackage{natbib}
\usepackage{graphicx}
\usepackage{subfigure}
\usepackage{textcomp}
\usepackage{xcolor}
\usepackage{algpseudocode}
\usepackage{algorithm}
\usepackage{url}
\usepackage{makecell}
\usepackage{bbding}

\usepackage{graphicx}
\usepackage{tabularx}
\usepackage{amsmath,xparse,mleftright}
\usepackage[export]{adjustbox}
\usepackage{balance}
\usepackage{tcolorbox}
\usepackage{tabularray}

\usepackage{wrapfig}
\usepackage{booktabs}
\usepackage{hyperref}
\usepackage{enumitem}
\usepackage{xcolor}
\usepackage{pifont}
\usepackage{xspace}
\usepackage{wrapfig}
\usepackage{setspace}
\usepackage{tabularx,ragged2e}
\usepackage{bm}

\newcommand{\eg}{\emph{e.g.},\xspace}
\newcommand{\ie}{\emph{i.e.},\xspace}
\newcommand{\etal}{\emph{et al.},\xspace}
\newcommand{\etc}{\emph{etc.}\xspace}
\newcommand{\name}{\textsc{XoT}\xspace}
\iclrfinalcopy{\iclrfinaltrue}
\usepackage{amsmath,amsfonts,bm}

\def\eqref#1{equation~\ref{#1}}

\def\1{\bm{1}}

\DeclareMathAlphabet{\mathsfit}{\encodingdefault}{\sfdefault}{m}{sl}
\SetMathAlphabet{\mathsfit}{bold}{\encodingdefault}{\sfdefault}{bx}{n}

\newcommand{\htick}{\raisebox{\ups\height}{\includegraphics[width=1em]{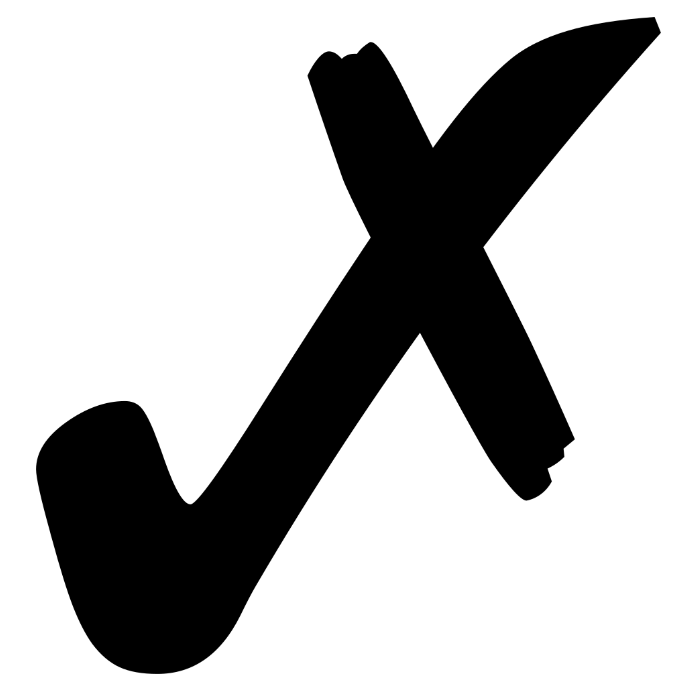}}}%
\newcommand{\Itriangle}{\raisebox{\ups\height}{\includegraphics[width=1em]{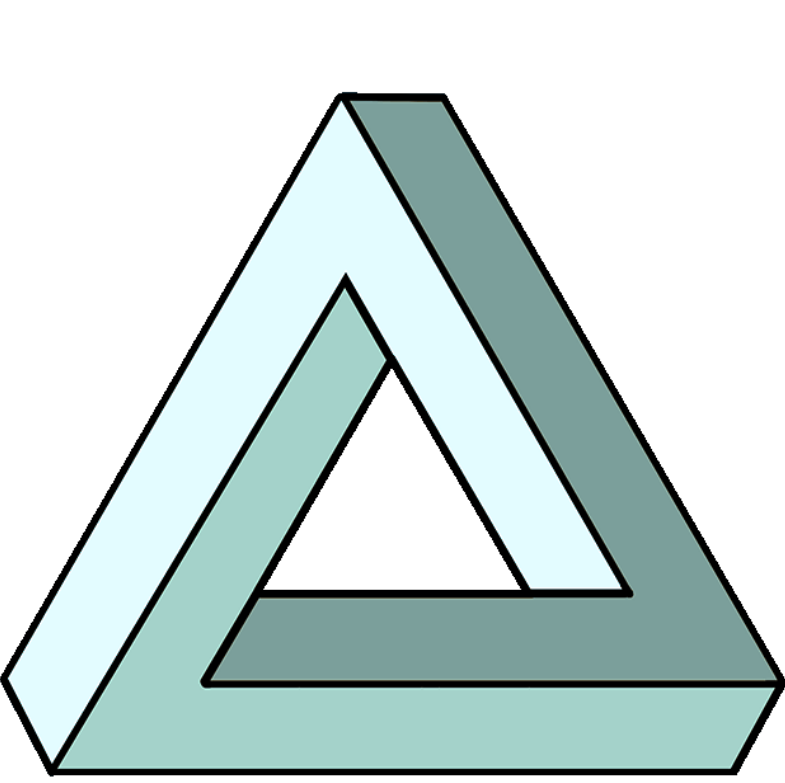}}}%
\newcommand{\ItriangleB}{\raisebox{\ups\height}{\includegraphics[width=1.em]{figure/Itriangle.pdf}}}%
\newcommand{\cube}{\raisebox{\ups\height}{\includegraphics[width=1.5em]{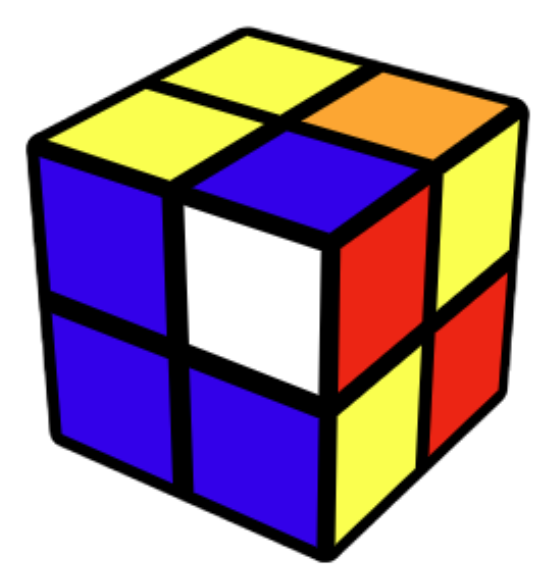}}}%
\newcommand{\cubes}{\raisebox{\ups\height}{\includegraphics[width=1.5em]{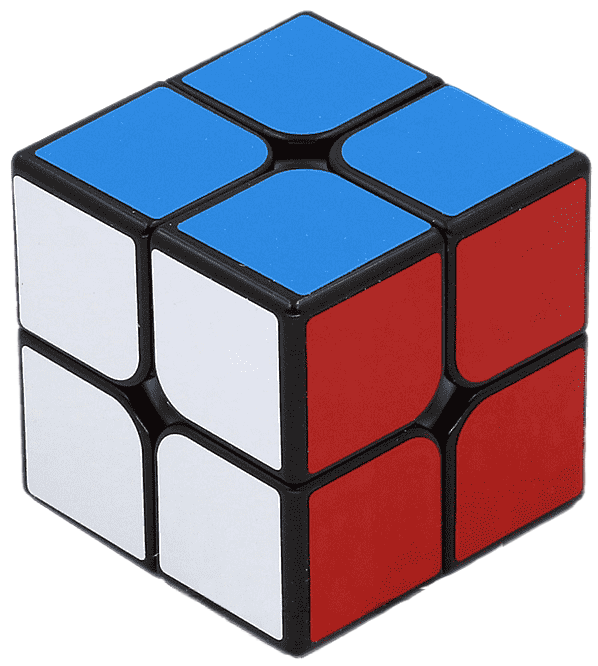}}}%
\newcommand{\nps}{\raisebox{\ups\height}{\includegraphics[width=1.7em]{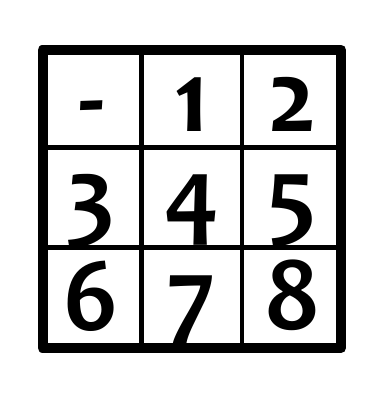}}}%
\newcommand{\npns}{\raisebox{\ups\height}{\includegraphics[width=1.7em]{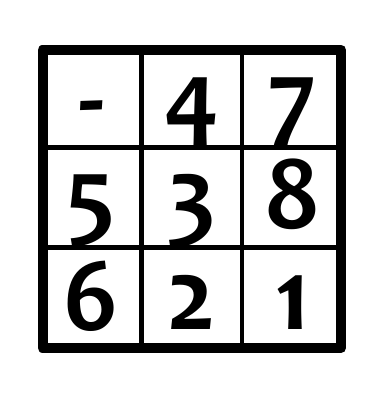}}}%
\newcolumntype{C}[1]{>{\centering\let\newline\\\arraybackslash\hspace{0pt}}m{#1}}
\newcolumntype{L}[1]{>{\raggedright\let\newline\\\arraybackslash\hspace{0pt}}m{#1}}

\algnewcommand{\Inputs}[1]{%
  \State \textbf{Inputs:}
  \Statex \hspace*{\algorithmicindent}\parbox[t]{.8\linewidth}{\raggedright #1}
}
\algnewcommand{\Initialize}[1]{%
  \State \textbf{Initialise:}
  \Statex \hspace*{\algorithmicindent}\parbox[t]{.8\linewidth}{\raggedright #1}
}

\def\ups{-0}
\def\cwidth{3.35cm}

\title{Everything of Thoughts \ItriangleB: Defying the Law of Penrose Triangle for Thought Generation}

\author{Ruomeng Ding\thanks{This work was completed during her internship at Microsoft Research Asia.},$^{1, 2}$ Chaoyun Zhang$^{1}$, Lu Wang$^{1}$, Yong Xu$^{1}$,  Minghua Ma$^{1}$, \\
\textbf{Wei Zhang$^{3}$, Si Qin$^{1}$, Saravan Rajmohan$^{1}$, Qingwei Lin$^{1}$ \& Dongmei Zhang$^{1}$}\\
$^{1}$Microsoft\\
$^{2}$Georgia Institute of Technology\\
$^{3}$East China Normal University
}

\begin{document}

\maketitle

\begin{abstract}
Recent advancements in Large Language Models (LLMs) have revolutionized decision-making by breaking down complex problems into more manageable language sequences referred to as ``thoughts''. An effective thought design should consider three key perspectives: performance, efficiency, and flexibility. However, existing thought can at most exhibit two of these attributes. To address these limitations, we introduce a novel thought prompting approach called ``Everything of Thoughts'' (\name) to defy the law of ``Penrose triangle \Itriangle'' of existing thought paradigms. \name leverages pretrained reinforcement learning and Monte Carlo Tree Search (MCTS) to incorporate external domain knowledge and planning capability into thoughts, thereby enhancing LLMs' capabilities and enabling them to generalize to unseen problems efficiently. Through the utilization of the MCTS-LLM collaborative thought revision framework, this approach autonomously produces high-quality comprehensive cognitive mappings with minimal LLM interactions. Additionally, \name empowers LLMs to engage in unconstrained thinking, allowing for flexible cognitive mappings for problems with multiple solutions.

We evaluate \name on several challenging problem-solving tasks, including Game of 24, 8-Puzzle, and Pocket Cube. Our results demonstrate that \name significantly outperforms existing approaches in various dimensions, showcasing its remarkable proficiency in addressing complex problems across diverse domains. The code and dataset to reproduce the results in the paper are available at \url{https://github.com/microsoft/Everything-of-Thoughts-XoT-}.
\end{abstract}

\section{Introduction}
\begin{wraptable}{r}{8cm}
\centering
\vspace*{-2em}
\caption{Comparisons of different prompting paradigms.\label{tab:compare}}
\begin{tabular}{lccc}
\hline
\textbf{Paradigm} & \textbf{Performance}                                                            & \textbf{Efficiency}                                                               & \textbf{Flexibility}                                                            \\ \hline
IO               & \XSolidBrush & \Checkmark                                                             & \XSolidBrush                                                           \\
CoT               & \htick & \Checkmark                                                             & \XSolidBrush                                                           \\
CoT-SC            & \htick & \htick & \XSolidBrush                                                           \\
ToT               & \Checkmark                                                             & \XSolidBrush                                                           & \htick \\
GoT               & \Checkmark                                                             & \XSolidBrush                                                           & \Checkmark                                                             \\ \hline
\textbf{\name}             & \Checkmark                                                             & \Checkmark                                                             & \Checkmark                                                             \\ \hline
\end{tabular}
\vspace*{-0.5em}
\end{wraptable}
Recent advancements in Large Language Models (LLMs) have greatly advanced problem solving in diverse domains such as mathematical reasoning \cite{frieder2023mathematical}, knowledge reasoning \cite{omar2023chatgpt}, root cause analysis \cite{chen2023empowering} and causal inference \cite{kiciman2023causal}, \etc. This progress can be largely attributed to the technique of decomposing intricate problems into smaller language sequences referred to as ``thoughts''. Through a step-by-step inference process involving the use of prompts, each thought functions as an intermediate stage, contributing to the simplification of tackling complex problems to fulfill the problem's ultimate objective. 

Effective design of thought steps toward complex problem-solving and reasoning, whether for humans or LLMs, should prioritize three crucial aspects, namely:
\begin{itemize}[leftmargin=*]
    \item \textbf{Performance.} Performance is the accuracy of the solution to a problem, including the precision of each thought at intermediate stages. This metric holds paramount importance for problem-solving.
    \item \textbf{Efficiency.} Efficiency relates to the number of LLM inference calls required to solve a single problem. Minimizing this aspect is crucial due to the high computational cost associated with LLM inference, thereby reducing the overall number of cost.
    \item \textbf{Flexibility.} 
    Flexibility in thought topology refers to the diverse structures that can be employed by LLMs when organizing thoughts for problem-solving. These structures may include chains, trees, or even graphs, mirroring human thought processes. Enabling more flexible thought structures enhances the capacity of LLMs for divergent and creative thinking, which is particularly advantageous in addressing complex problems, especially those with multiple potential solutions.
\end{itemize}
There exist several thought generation paradigms, such as Chain-of-Thought (CoT) \cite{wei2022chain}, Tree-of-Thought (ToT) \cite{yao2023tree}, and Graph-of-Thought (GoT) \cite{besta2023graph}, \etc. However, these paradigms each have their limitations and cannot simultaneously achieve all the three desired attributes, as illustrated in Table~\ref{tab:compare}. Specifically, direct Input-Output (IO) prompting is suitable primarily for simple problem-solving scenarios with single-step processes, lacking both in performance and flexibility. CoT and self-consistency CoT (CoT-SC) enable step-by-step problem solving, resulting in modest performance improvements, but they are confined to linear thought structures, limiting their flexibility. In contrast, ToT and GoT permit more versatile thought topologies, accommodating tree-like or graph-like structures. However, these paradigms require the evaluation of intermediate thought steps through LLM itself, incurring significant computational costs and inefficiencies due to multiple LLM calls. These paradigms are constrained by a law analogous to the ``Penrose triangle \Itriangle'', wherein they can achieve a maximum of two out of the three attributes, and none of them can simultaneously attain all three.

We propose a novel solution called ``Everything of Thoughts'' (\name) to address the limitations of conventional thought frameworks, enhancing essential attributes of thought generation, including performance, efficiency, and flexibility for LLM inference.\footnote{We named it ``Everything of Thoughts'' to signify its three comprehensive thought generation capabilities.} \name leverages reinforcement learning (RL) \cite{li2017deep} and Monte Carlo Tree Search (MCTS) \cite{silver2017mastering}, in conjunction with lightweight policy and value networks, to pretrain on specific tasks for thought searching and subsequently generalize to new problems. This pretraining effectively integrates external domain knowledge and planning capability into the ``thoughts'' provided to LLMs, expanding their problem-solving capabilities, and thereby significantly improving \textbf{Performance}. Once trained, \name efficiently performs thought searching using MCTS with cost-effective policy and value networks for exploration and autonomously generates complete cognitive mappings for LLMs. It then employs a \textbf{MCTS-LLM collaborative thought revision process} to further improve the thought quality while minimizing LLM interactions. This eliminates the need for LLMs to explore and evaluate thoughts themselves, as required by ToT and GoT, enhancing \name's \textbf{Efficiency}. Furthermore, MCTS demonstrates remarkable \textbf{Flexibility} as it can explore various thought topologies, including graph structures akin to those employed in human mind mapping processes \cite{faste2012untapped, jamieson2012using}. This enables diverse and creative thinking for LLMs, making it particularly valuable when dealing with complex thought structures or tasks featuring multiple potential solutions. By concurrently achieving superior performance, efficiency, and flexibility, \name challenges the constraints posed by the ``Penrose triangle \Itriangle'' law, significantly surpassing the capabilities of other thought generation paradigms.

We comprehensively evaluate \name across a diverse range of challenging problem-solving tasks, namely Game of 24, 8-Puzzle, and Pocket Cube. Our experimental results consistently showcase \name's superior performance, and its capacity to provide multiple solutions to problems efficiently with just a few LLM calls. These findings establish \name as an effective thought generation approach, paving the way for new avenues in LLMs' problem-solving capabilities.

\vspace*{-0.5em}

\section{Background}

\noindent \textbf{Thought for LLMs.}
Addressing complex problems often entails breaking down the overarching objective into multiple intermediary steps. 
The outcomes or cognitive processes associated with each step are thoughts, which can be expressed as linguistic prompt sequences for LLMs to facilitate problem-solving. Structures of these thought may take various forms, including linear chains, hierarchical trees, or interconnected graphs, depending on how the thoughts are organized to advance towards a solution. 



\begin{figure}[t]
\centering
\includegraphics[width=1\columnwidth]{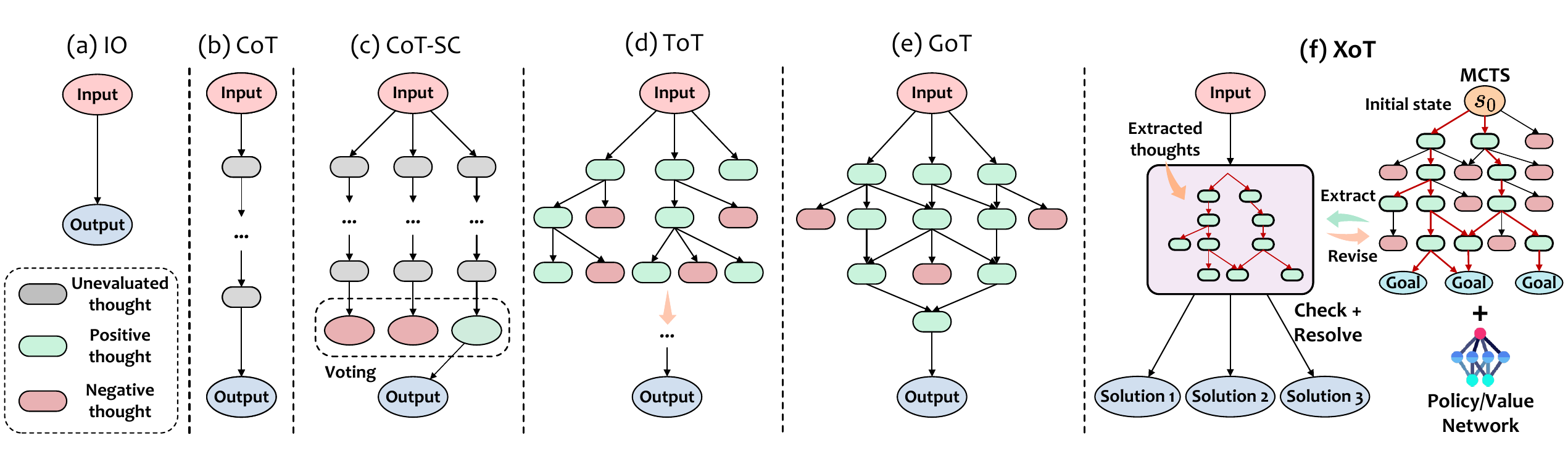}
\vspace*{-2.5em}
\caption{Comparison of \name versus other prompting paradigms.
\label{fig:compare}}
\end{figure}

\noindent \textbf{Input-Output (IO) Prompting (Fig.~\ref{fig:compare} (a)).} 
The IO method is the most straightforward approach to instruct LLMs to address a problem without the provision of any intermediate thought processes.

\noindent \textbf{Chain-of-thought (CoT) \cite{wei2022chain} (Fig.~\ref{fig:compare} (b)).}
CoT decomposes problem-solving into a sequential chain of thoughts, allowing LLMs to approach complex problems step by step.

\noindent \textbf{Self-consistency CoT (CoT-SC) \cite{wang2023selfconsistency} (Fig.~\ref{fig:compare} (c)).} 
CoT-SC employs multiple instances of the CoT to generate multiple outputs from LLMs. It selects the the best results from multiple LLM outputs, offering  more robust and consistent inference compared to the vanilla CoT.

\noindent \textbf{Tree-of-thought (ToT) \cite{yao2023tree} (Fig.~\ref{fig:compare} (d)).} 
ToT organizes thoughts in a tree-like structure and utilizes search algorithms (\eg Breadth-First Search, Depth-First Search) to expand the tree in pursuit of an optimal solution. However, thought evaluation in ToT relies on LLMs themselves, necessitating multiple costly and inefficient LLM inference calls.

\noindent \textbf{Graph-of-thought (GoT) \cite{besta2023graph} (Fig.~\ref{fig:compare} (e)).} 
GoT extends the ToT approach by enabling the generation of graph-like thought structures through thought aggregation and refinement during intermediate search phases. Although this method permits more flexible thought structures, it still demands multiple LLM inference calls for evaluation, incurring significant computational costs.

\vspace*{-0.5em}

\section{\name: Everything of Thoughts}

\name serves as an LLM-MCTS collaborative framework designed to enhance the thought generation process, thereby assisting LLMs in resolving complex problems. It leverages MCTS for proficient and efficient thought exploration while harnessing the capabilities of LLMs to refine and amend the thoughts derived from MCTS. This synergistic interaction creates a mutually beneficial arrangement, ultimately enabling the successful resolution of intricate problems characterized by high levels of performance, efficiency, and flexibility.

\subsection{\name in a Nutshell}
We present an overview of the architecture of \name in Fig.~\ref{fig:compare} \textbf{(f)}. \name comprises two key components: \emph{(i)} a MCTS module guided by policy/value networks; and \emph{(ii)} an LLM solver for thought revision and inference. The MCTS and policy/value networks need to be trained and then generalize to the inference process.

During the training phase, MCTS is harnessed to explore potential thought structures for a specific task through simulated scenarios. This process entails the recording of states, values, and the visitation frequencies of thought nodes in each simulation. These recorded data are subsequently employed to iteratively train the policy and value estimation model, enabling it to assimilate domain knowledge and comprehend the world model. 

Once trained, the estimated policy and value are utilized to guide the MCTS to systematically search for a thought trajectory provided to aid LLMs in problem-solving. Note that thoughts extracted only play a supporting role, assisting LLMs in gathering knowledge from external sources and improving its planning capability . These thoughts do not provide LLMs with definitive or error-free answers, as they may contain inaccuracies or suboptimal solutions. LLMs are responsible for review and refining these thoughts when they seem erroneous or require adjustments. They continue MCTS the search process if needed and eventually formulate the final answers by integrating these external thoughts with their internal knowledge.





\vspace*{-0.5em}
\subsection{Thought Searching Formulation}
The fundamental objective of employing the thought generation paradigm for LLMs is to identify the optimal decomposition of a complex problem into several manageable sub-steps. Each sub-step aims to alter the current status of the problem, eventually culminating in the successful resolution of the overarching problem. This approach, as seen in ToT and GoT, hinges on well-defined state transitions and clear final objectives. Consequently, it is natural to conceptualize the thought-searching process as a Markov Decision Process (MDP) \cite{puterman1990markov}, in which:

\begin{itemize}[leftmargin=*]
    \item \textbf{State $s_t$:} Represents the current status of the problem. The initial state $s_0$ corresponds to the original problem, while intermediate states are characterized by either decomposed sub-problems or the results stemming from their resolution.
    \item \textbf{Action $a_t$:} Signifies the one-step solution or action associated with tackling a problem, leading to a transition to a new state, by incorporating their outcomes.
    \item \textbf{Reward $r$:} Reflects the comprehensive evaluation of the solution to the original problem, assessing whether it has been effectively resolved through the process of problem decomposition.
    \item \textbf{Thought $\tau$:} A one-step thought is a combination of one-step state and action, \ie $\tau = \{s, a\}$. This formulation naturally encapsulates the process of decomposing a complex problem into multiple sub-tasks, each accompanied by their respective outcomes.
\end{itemize}
The detailed definitions of state, action, reward and thought for each task are shown in Table~\ref{tab:compare}.
The generation of complete thoughts $\mathcal{T} = \{\tau_1, \cdots, \tau_N\}$, can be construed as the endeavor to discover a thought trajectory to maximize the accumulated reward to address the overall problem. 

\vspace*{-0.5em}
\subsection{Thoughts Searching with MCTS}
\begin{figure}[t]
\centering
\includegraphics[width=1\columnwidth]{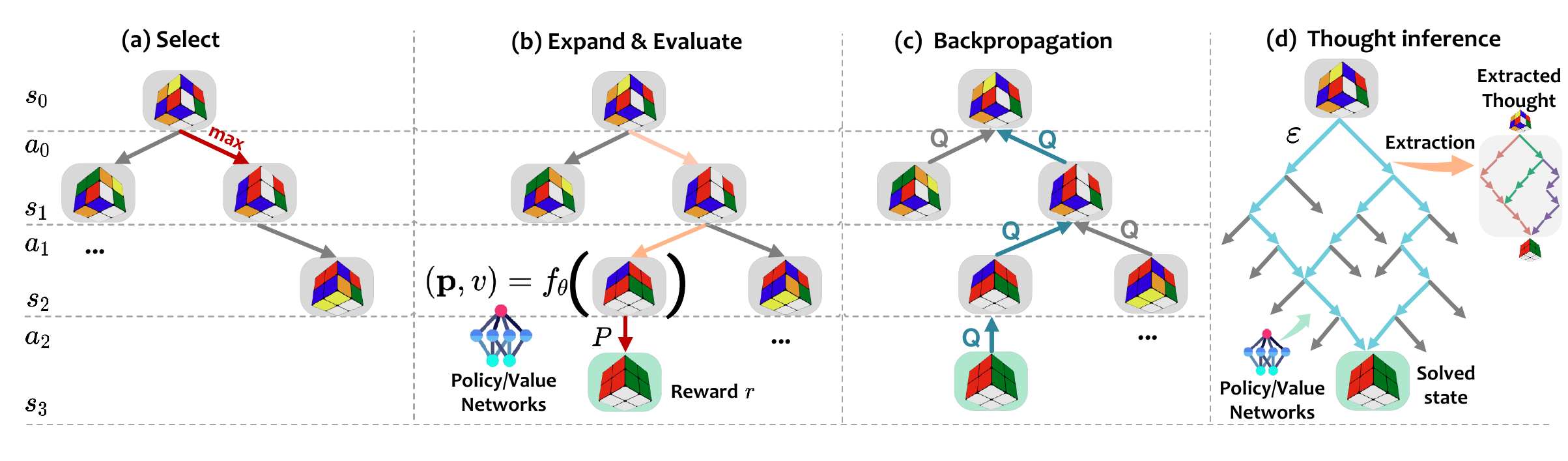}
\vspace*{-2.5em}
\caption{An illustration of iterative phases in MCTS for thought searching (\textbf{(a)-(c)}) and thought inference in problem resolution \textbf{(d)}.
\label{fig:mcts}}
\end{figure}

The formulation above naturally aligns the thought within LLM as a state-action pair. This approach facilitates the effective exploration of its optimal trajectory using a combination of MCTS and RL. This adheres to an iterative simulation cycle that encompasses three key phases: selection, expansion \& evaluation, and backpropagation. It heavily depends on the utilization of neural networks $f_\theta$, which simultaneously estimate the value and action probability for a given state $s_t$. The aim is to reduce the number of rollouts and accelerate the search process, similar to the approach employed in AlphaGo Zero \cite{silver2017mastering}. We provide a visual representation of an iteration of the MCTS in Fig.~\ref{fig:mcts} \textbf{(a)-(c)} by taking Pocket Cube as an example and detail each process below.


\noindent \textbf{Selection.} In the selection phase, the algorithm initiates at the root node and proceeds to choose an action $a^*$ from the available set $\mathcal{A}(s)$ for single-step thought generation in the current state $s$. This process continues until a leaf node within the current tree is reached. The selection is guided by the PUCT algorithm \cite{rosin2011multi}, aiming to maximize the Upper Confidence Bound (UCB) \cite{garivier2011upper}, as follows:
\begin{align}
    a^* = \arg \max _{a\in \mathcal{A}(s)} \left [Q(s,a) + w \cdot P_\theta(s, a)\sqrt{\frac{N(s)}{1+N(s, a)} }   \right ].
    \label{eq:UCB}
\end{align}
Here, $Q(s,a)$ denotes the Q-value of a state-action pair $(s,a)$, which estimates the quality of a particular action in a given state. The higher the Q-value, the better the action is considered to be.
$P_\theta(s, a)$ denotes the predicted prior probability of selecting action $a$ given the state $s$ obtained from a neural network $f_\theta$, and $N(s, a)$ represents the count of times action $a$ has been chosen in state $s$. The parameter $w$ controls the trade-off between exploration and exploitation. The selection process will continue until an unexplored node is encountered.


\noindent \textbf{Evaluation and Expansion.} 
Upon reaching a previously unselected leaf node, we expand to the state $s$ for the next step for new thought exploration. This expansion involves the evaluation of its value and action probability on the state, which are modeled by neural networks parameterized by $\theta$, \ie $(P_\theta(s), v_\theta(s)) = f_\theta(s)$. Here $P_\theta(s)$ is the prior probabilities for all actions on $s$, and $v_\theta(s)$ denotes its predicted state value. These two values are retained and stored for backup purposes, and state $s$ is masked as ``visited''.

\noindent \textbf{Backpropagation.}
Following the expansion of a leaf node in the above phases, which could be either an unexplored or terminal state, the algorithm proceeds to update all the $Q(s, a)$ values via backpropagation. For unexplored nodes, this update involves computing the mean of its estimated value $v_\theta$, while for terminated nodes, it's based on the true reward $r$. These updates occur as information is backpropagated along the trajectory to subsequent nodes. Additionally, the visit count for each state-action pair is also incremented as follows: $N(s, a) = N(s, a) + 1$. 

A simulation is completed after a sequence of selection, evaluation, expansion, and backpropagation steps. After conducting multiple simulations, we proceed to the next step by selecting an action at state $s$ using a probability distribution defined as $\varepsilon_a \propto N(s, a)^{1/\gamma}$, where $\gamma$ is a temperature constant that regulates the level of exploration.

\noindent \textbf{Policy and Value Networks Training.}
The simulations described above allow us to compile a dataset for each sample state $s$ containing $(s, \bm{\varepsilon}(s), v(s))$, where $\bm{\varepsilon}(s) = \{\varepsilon_a \mid a \in \mathcal{A}(s)\}$, and $v(s)$ represents the ground truth value obtained by accumulating rewards along the trajectory starting from state $s$. Subsequently, we can train a combined policy and value network $f_\theta$ to minimize the discrepancy between the predicted value $v_\theta(s)$ and the actual value $v(s),$ while also maximizing the alignment between the action probabilities produced by the neural network $P_\theta(s)$ and the search probabilities $\bm{\varepsilon}(s)$. This can be achieved by minimizing the following loss function:
\begin{align}
    \mathcal{L} = (v(s) - v_\theta(s))^2 + \bm{\varepsilon}(s)^T \log P_\theta(s)).
\end{align}
This training iterates alongside the simulation process to continually enhance the performance of $f_\theta$, resulting in progressive improvements in thought searching capabilities.

\subsection{Thought Inference with MCTS}
Once trained, we utilize the $f_\theta$ to guide the MCTS in generating a thought for a new problem, which assists the LLM in solving it. Specifically, MCTS is utilized to perform $K$ simulations aimed at thought searching and problem-solving, as illustrated in Fig.\ref{fig:mcts} \textbf{(d)}. In each simulation, $f_\theta$ is employed to guide the MCTS in its search for a thought trajectory. Throughout the training process, $f_\theta$ incorporates external information related to the state and action quality. This information helps LLMs understand the world model, enhancing their long-term reasoning and planning abilities, which are areas they may not excel in \cite{stechly2023gpt, valmeekam2023can}, thereby ensuring the \emph{performance} of thought generation. Once the simulation concludes, we record the visiting count $N(s, a)$ and the thought trajectory is obtained based on the number of solutions required:
\begin{itemize}[leftmargin=*]
    \item \textbf{Single solution.} starting from each state $s$, the action with the highest visiting count $N(s, a)$ is selected.
    \item \textbf{Multiple solution.} we sample $M$ thought trajectories following the probability distribution $\varepsilon_a \propto N(s, a)$ and remove duplicates.
\end{itemize}
This results in one or multiple thought trajectories $\mathcal{T}^*$ that consist of a sequence of state-action pairs for problem-solving. The trajectories for multi-solution problems may intertwine and converge at the same goal state, resulting in a graph-like thought structure. This demonstrates that \name is capable of generating thought structures with \emph{flexibility}. These trajectories are then transformed into text sequences that are concatenated to form a prompt sequence provided to LLMs. Note that the thought trajectory is concatenated into a single prompt, even in the case of problems with multiple solutions. Therefore, we only require a single LLM inference call at this stage. Given that the $f_\theta$ network is relatively lightweight, this ensures the \emph{efficiency} of \name.

\begin{figure}[t]
\centering
\includegraphics[width=1\columnwidth]{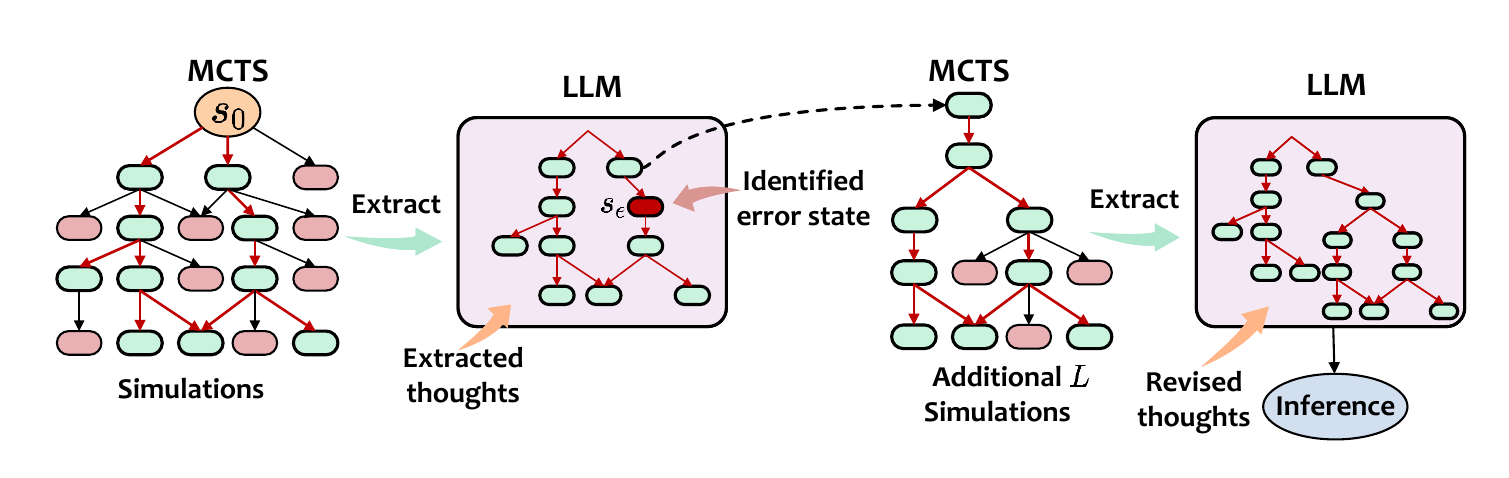}
\vspace*{-3.em}
\caption{An illustration of thought revision process in \name.
\label{fig:revision}}
\vspace*{-1.5em}
\end{figure}

\textbf{Thought-to-Prompt Parsing.}
Once the thought trajectories $\mathcal{T}^*$ are extracted from MCTS, we convert them into a textual format necessary for LLM inference. In this conversion process, we transform both the state and action at each step of the thought, \ie $\tau = \{s, a\}$ in $\mathcal{T}^*$, into text. This conversion aims to provide a comprehensive state transition, facilitating LLMs in better understanding the task step by step. In the case of multi-solution scenarios, multiple trajectories are concatenated. This format remains consistent across all baselines, and the resulting prompt text is then fed to LLMs for inference or thought revision.

\textbf{Thought Revision.}
It is important to acknowledge that that MCTS may not always provide the globally optimal thought trajectory to directly solve the problem flawlessly. Therefore, the thoughts extracted from MCTS serve as a reference thinking process for the problem, aiding LLMs in a \emph{supportive} capacity. The LLMs will leverage their internal knowledge to review the extracted thought, identify errors in the thought trajectory, and then ground its knowledge in collaboration with the MCTS to revise and refine the thought. In this context, LLM plays a role akin to a participant in the collaborative framework, guiding MCTS to enhance its performance.

The revision process is iterative in nature, as shown in Fig.~\ref{fig:revision}. Initially, upon obtaining the extracted thought, we instruct the LLM to detect any errors in the thought generated by MCTS using its internal knowledge. If the LLM identifies an error, it results in an error state denoted as $s_e$ within the thought. If no error is found, the thought remains unchanged. Starting from the parent state of $s_e$, MCTS conducts an additional set of $L$ simulations, ultimately yielding a revised thought for the LLM. In scenarios involving multiple solutions, each solution undergoes this process individually. Upon the completion of the revision, we supply the LLMs with the revised thoughts for problem-solving. The revision process can be repeated several times to enhance the reliability of the answer. This collaborative MCTS-LLM framework nurtures a mutually beneficial process for both components, ultimately contributing to the overall performance of problem-solving. Since LLMs are solely utilized for identifying errors during the revision process with only one call, the efficiency of \name is effectively maintained.

The collaborative revision framework harnesses the strengths of both MCTS and LLMs. MCTS efficiently and flexibly generates candidate thoughts for LLMs through simulations, while LLMs use their internal knowledge to revise and ground these thoughts within the MCTS framework, effectively turning MCTS into a world model for LLMs. This process ensures the generation of high-quality thoughts for problem-solving.

\vspace*{-0.5em}

\section{Experiment}
\begin{table}[t]
\caption{An overview of tasks employed in this study. \label{tab:tasks}}
\begin{tabular}{l|C{\cwidth}C{\cwidth}C{\cwidth}}
\hline
                       & \textbf{\textbf{Game of 24}}                                                                   & \textbf{\textbf{8-Puzzle}}                                                                         & \textbf{\textbf{Pocket Cube}}                                                                          \\ \hline
\textbf{Objective}     & Use four numbers on playing cards to make the number 24 through $+$, $-$, $\times$, or $\div$. & Rearrange the tiles in the $3\times 3$ puzzle from an scrambled state to a goal state \nps.        & Rotating the faces of a $2\times 2$ pocket cube until each face of the cube is a uniform color \cubes. \\ \hline
\textbf{Input}         & 4 numbers ranging from 1 to 13, \eg (4, 6, 10, 10).                                            & A scrambled $3 \times 3$ digital puzzle, \eg \npns.                                                & A scrambled $2 \times 2$ pocket cube, \eg \cube. Colors represented as numbers for LLMs.               \\ \hline
\textbf{Output}         & An equation to reach 24, \eg $4\times 6+10-10=24$.                                             & The slide sequence of the ``-'' tile, \eg (Up, Down, Left, Right $\cdots$).                        & The rotation move sequence of the cube, \eg (F, R2, U' $\cdots$).           \\ \hline
\textbf{Thought}       & 3 intermediate equations.                                                                      & The step-by-step sliding, and the puzzle state after the move.                                     & The step-by-step rotation, and the cube state after the move.                                          \\ \hline
\textbf{State}     & The remaining 1-4 numbers.                                                     & The current number layout of the puzzle.                                                           & Colors of each face of the pocket cube.                                                                \\ \hline
\textbf{Action}    & Picking two number and a operation to compose an equation.                                     & The one-step moving action of the ``-'' tile.                                                      & The one-step rotation action of cube.                                                                  \\ \hline
\textbf{Reward}    & 1 if the number of the final number is equal to 24 otherwise -1.                               & The negative  minimum step on solving the current puzzle state toward the goal state. & The negative minimum moving step on solving current cube state toward the goal state.    \\ \hline
\end{tabular}
\vspace*{-0.5em}
\end{table}

We conduct an extensive evaluation of our \name approach in comparison to several baseline methods across three challenging tasks: the Game of 24, the 8-Puzzle (with a $3\times 3$ grid), and the $2\times 2$ Pocket Cube. An overview of these tasks is provided in Table~\ref{tab:tasks}. These tasks are characterized by their complexity, requiring multiple steps for completion and potentially having multiple solutions. To assess the effectiveness of \name, we compare it against IO, CoT, CoT-SC, ToT, GoT, and single MCTS without LLMs for inference and revision. We also finetune LLaMA-2-13B \cite{touvron2023llama} for comparison, using the same training data and ground truth labels. The setup of LLaMA-2-13B can be found in Appendix~\ref{appendix.baseline}. We employ both GPT-3.5 \cite{ouyang2022training} and GPT-4 \cite{openai2023gpt4} for these evaluations. Note that  \texttt{temperature} and \texttt{top\_p} are set to 0.0 for all LLM invoked. We further conduct ablation study to assess the impact of thought revisions, the revision success rate, and the sensitivity to the completeness of the provided thoughts, presented in Section~\ref{sec.ablation}. We conduct case study in Multi-Solution Scenarios in Section~\ref{sec.case_study} to illustrate the thought structures. The computational training costs of MCTS are discussed in Appendix~\ref{appendix.training_cost}. The discussion on generalizing \name to other NLP tasks, such as Document Merging \cite{besta2023graph}, can be found in Appendix~\ref{appendix.nlp}.

\textbf{Policy/Value Networks Configurations}. The policy and value networks in our model utilize a shared multi-layer perceptron (MLP) architecture with two layers and hidden units arranged as (128, 256). Two heads connected to the MLP are responsible for predicting $v_\theta(s)$ and $P_\theta(s)$ separately. The total number of parameters in the Policy/Value Network for all three tasks is approximately $10^6$. This design results in a considerably smaller model compared to LLM, making it much more efficient. We train this model through three iterations, with each iteration comprising 10 self-play episodes for MCTS.


\textbf{Evaluation Metric.} For each task, we assess the accuracy of each approach on the test set. Additionally, we track the number of LLM invocations required for all approaches to solve a problem, as well as the number of times $f_\theta$ is invoked in the case of \name. It's important to note that $f_\theta$ is a considerably smaller model compared to LLMs. In the context of multi-solution scenarios, accuracy is computed as the percentage of problems for which any of the answers provided by each approach is correct. Multi-solution Accuracy (\textbf{MultiAcc}) is calculated as the average percentage of correctness across all solutions offered. Furthermore, we capture the total count of distinct solutions provided by each approach, regardless of their correctness, represented as \textbf{\#Sol}. Note that we set the maximum solution number to 3 for all problems in multi-solution scenarios. In Table~\ref{tab:game24.with_laststep} to Table~\ref{tab:cube.multi}, the number of thought \textbf{revision} is denoted by \textbf{r}.

\begin{table}[t]
\small
\centering
\caption{Performance comparison on Game of 24.\label{tab:game24.with_laststep}}
\resizebox{0.9\width}{!}{
\begin{tabular}{l|ccc|ccc} 
\hline
\multirow{2}{*}{\textbf{Model}} & \multicolumn{3}{c|}{\textbf{GPT-3.5}}                          & \multicolumn{3}{c}{\textbf{GPT-4}}                              \\ 
\cline{2-7}
& \multicolumn{1}{p{1.2cm}}{\centering\scriptsize\textbf{Acc. [\%]}} & \multicolumn{1}{p{1.2cm}}{\centering\scriptsize\textbf{LLM invoked}} & \multicolumn{1}{p{1.2cm}|}{\centering\scriptsize\textbf{$f_\theta$ \\invoked}} & \multicolumn{1}{p{1.2cm}}{\centering\scriptsize\textbf{Acc. [\%]}} & \multicolumn{1}{p{1.2cm}}{\centering\scriptsize\textbf{LLM invoked}} & \multicolumn{1}{p{1.2cm}}{\centering\scriptsize\textbf{$f_\theta$ \\invoked}}   \\ 
\hline
IO                              & 6.57               & 1.00                 & -                  & 10.22              & 1.00                 & -                   \\
CoT                             & 2.19               & 1.00                 & -                  & 4.38               & 1.00                 & -                   \\
CoT-SC                  & 2.19               & 10.00                & -                  & 4.38             & 10.00                & -                   \\
ToT (b=1)                       & 5.84               & 22.11                & -                  & 34.31              & 23.50                & -                   \\
ToT (b=3)                       & 10.22              & 43.96                & -                  & 60.58              & 39.83                & -                   \\
GoT (k=1)                       & 2.92               & 7.00                 & -                  & 10.95              & 7.00                 & -                   \\ 
LLaMA-2-13B                        & 2.19               & -                 & -                  & 2.19              & -                 & -                   \\ 
MCTS                        & 62.77               & -                 & -                  & 62.77              & -                 & -                   \\ 
\hline
\textbf{XoT (w/ 1 r)}               & \textbf{79.56}              & \textbf{1.39}                 & \textbf{92.15}              & \textbf{74.45}              & \textbf{1.38}                & \textbf{88.20}               \\
\textbf{XoT (w/ 2 r)}               & \textbf{88.32}              & \textbf{1.58}                 & \textbf{93.87}              & \textbf{83.94}              & \textbf{1.57}                & \textbf{89.63}               \\
\textbf{XoT (w/ 3 r)}               & \textbf{90.51}              & \textbf{1.72}                 & \textbf{95.94}              & \textbf{85.40}              & \textbf{1.78}                & \textbf{92.48}               \\
\hline
\end{tabular}}
\end{table}

\begin{table}[t]
\small
\centering
\caption{Performance comparison on Game of 24 in the multi-solution scenario.\label{tab:game24.multi}}
\resizebox{0.9\textwidth}{!}{
\begin{tabular}{l|cccc|cccc} 
\hline
\multirow{2}{*}{\textbf{Model}} & \multicolumn{4}{c|}{\textbf{GPT-3.5}}                                                                                                      & \multicolumn{4}{c}{\textbf{GPT-4}}                                                                                                          \\ 
\cline{2-9}
                                & \begin{tabular}[c]{@{}c@{}}\textbf{Multi }\\\textbf{Acc.}\end{tabular}  & \textbf{\#Sol} & \begin{tabular}[c]{@{}c@{}}\textbf{LLM }\\\textbf{invoked}\end{tabular} & \begin{tabular}[c]{@{}c@{}}\textbf{$f_\theta$}\\\textbf{invoked}\end{tabular}  & \begin{tabular}[c]{@{}c@{}}\textbf{Multi }\\\textbf{Acc.}\end{tabular} & \textbf{\#Sol} & \begin{tabular}[c]{@{}c@{}}\textbf{LLM }\\\textbf{invoked}\end{tabular} & \begin{tabular}[c]{@{}c@{}}\textbf{$f_\theta$}\\\textbf{invoked}\end{tabular}  \\ 
\hline
IO                              & 4.87              & 2.88           & 1.00                                                                    & -                           & 8.27              & 2.99           & 1.00                                                                    & -                            \\
CoT                             & 1.22              & 2.77           & 1.00                                                                    & -                           & 7.79              & 2.94           & 1.00                                                                    & -                            \\
CoT-SC                  & 1.70              & 2.76           & 10.00                                                                   & -                           & 8.03              & 2.99           & 10.00                                                                   & -                            \\
ToT (b=3)                       & 3.41              & 2.99           & 43.96                                                                   & -                           & 39.90             & 2.78           & 39.83                                                                   & -                            \\
GoT (k=3)                       & 8.03              & 1.93           & 7.00                                                                    & -                           & 10.46             & 1.39           & 7.00                                                                    & -                            \\ 
\hline
\textbf{XoT (w/ 1 r)}        & \textbf{62.90}    & \textbf{2.29}  & \textbf{3.51}                                                           & \textbf{116.34}             & \textbf{76.25}    & \textbf{2.36}  & \textbf{2.31}                                                           & \textbf{109.64}              \\
\hline
\end{tabular}}
\end{table}

\subsection{Game of 24}
The Game of 24 presents a arithmetic challenge wherein the goal is to employ four numbers within the range of 1 to 13, in conjunction with basic arithmetic operations, (\ie  $+$, $-$, $\times$, $\div$), to attain a final result of 24. This game may possess multiple valid solutions.

\subsubsection{Task Setup} 
We collect a dataset from \cite{4nums}, comprising 1,362 games ranked by human solving time, spanning a range of difficulty levels from easy to hard. For our testing phase, we randomly selected 137 games, ensuring coverage of various difficulty intervals. The remaining 1,225 problems were used to train the policy/value networks with MCTS. In the context of this task, as outlined in Table~\ref{tab:compare}, the thoughts refer to the three intermediate equations, while the state encompasses the available numbers (ranging from 1 to 4) for creating the equations. Actions involve the selection of two numbers and an operator to form an equation, and the reward is set to 1 if the final equation is both valid and results in the number 24, utilizing each of the input numbers exactly once, otherwise it is set to -1. Performance is measured by calculating the success rate across the 137 test games.

\subsubsection{Baselines \& \name Setup} 
The IO prompt is supported by five in-context examples. In the case of CoT, we augment each input-output pair by including three intermediate equations. As for ToT, we solicit one-step thought candidates from the LLM at each step, subsequently instructing the LLM to categorize each thought candidate for intermediate selection. For experimental comparison, we conduct experiments on both the top-1 candidate (with b=1) and the top-3 candidates (with b=3) being retained, where $b$ indicates the branches retained for exploration at each step. For GoT, we employ LLM to generate one-step thought candidates in the same manner as ToT, then we direct the LLM to select the top-1 thought from all candidates for merging the thoughts. We also examine a CoT-SC baseline, which derives the majority output from 10 CoT samples. For \name, we perform 200 simulations for each action taken, and this count is increased to 500 during the thought revision process.

In the multi-solution scenario, the IO, CoT, and CoT-SC prompts each include 5 examples, with each problem having 1 to 3 different solutions. For ToT, the top-3 candidates (with b=3) at the final step are considered as different solutions. Rather than keeping only the top-1 thought, GoT is instructed to select between 1 to 3 thoughts from all candidates at each step to generate a wider range of solutions. As for \name, after performing simulations on MCTS, we sample 500 thought trajectories as for exploration and remove deplicates. The top-3 thoughts with the highest counts are preserved.

\subsubsection{Results} 

Table~\ref{tab:game24.with_laststep} displays the overall performance of all methods on this task. Notably, \name consistently outperforms other baselines on both GPT-3.5 and GPT-4, achieving an accuracy of 79.56\% and 74.45\% respectively, with 1-time revision. However, after 3-time revision process, \name's accuracy substantially improves to 90.51\% and 85.40\% for GPT-3.5 and GPT-4 respectively. This underscores the impressive performance of \name, and demonstrates that the revision process significantly enhances performance, with only a limited increase in the utilization of LLM and $f_\theta$. Interestingly, the revision process in \name mitigates the performance gap attributable to the modeling ability in this task. As we observe that \name with GPT-3.5 achieves higher accuracy after revision compared to GPT-4.

Moreover, \name consistently outperforms the use of MCTS solely. The performance advantages exhibit growth with the number of revision iterations, underscoring the complementary roles of LLM and MCTS, emphasizing their joint necessity in achieving superior results. The fine-tuned LLaMA-2-13B is only successful on 2.19\% of the test data. This performance is lower than the IO method, indicating that the finetuning method is not be suitable for planning tasks like the Game of 24. The best-performing prompting baseline, ToT (b=3) on GPT-4, attains an accuracy of 60.58\%. However, it demands a substantial number of LLM invocations (39.83), which results in inefficiency. In contrast, \name only requires less than 1.8 calls with revision. Although \name requires some inference calls for $f_\theta$, the model is significantly less complex than LLM, making it a much more efficient approach.

Table~\ref{tab:game24.multi} presents the performance of different methods in the multi-solution scenario. Overall, \name remains the best-performing approach in terms of MultiAcc, significantly outperforming other baselines.  Although \name does not generate the most number of answers compared to other baselines, it generates more accurate answers, as its MultiAcc significantly outperforms other approaches. Notably, generating multiple solutions does not significantly increase \name's complexity, as it only requires 2.31 LLM calls with GPT-4 and around 100 calls for a smaller $f_\theta$, making it remain efficient. Overall, the remarkable performance of \name in the multi-solution scenario demonstrates its ability to generate complex thoughts.

\subsection{8-Puzzle}
The 8-Puzzle is a classic sliding puzzle game that consists of a $3\times 3$ grid with eight numbered tiles and one empty space denoted as ``-''. Its objective is to rearrange the tiles from a given initial configuration into a target configuration. The maximum number of steps necessary for the optimal solution of the 8-Puzzle is 31. This problem falls within the category of NP-complete problems \cite{ratner1986finding} and may have multiple solutions.

\subsubsection{Task Setup} 
We randomly generated 419 solvable 8-puzzle problems, with 300 instances allocated for training and 119 instances for testing. All generated problems are solvable within 9 steps. The action space encompasses four directions: [Up, Down, Left, Right]. Note that the legal action space for each problem state may vary due to the dynamic position of the empty space. As shown in Table~\ref{tab:compare}, the thoughts refer to the step-by-step move, and the puzzle state after the move.

\subsubsection{Baselines \& \name Setup} 
The IO prompt is extended with three in-context examples. In the CoT approach, each input-output pair is enriched by incorporating intermediate legal action sets, the current action, and the current state. In ToT, at each stage, a set of one-step thought candidates are derived from the LLM, from the current set of legal actions. We impose a maximum step limit of 9 since all generated problems can be solved within this range. The 8-puzzle's rules are conveyed through a system message, including detailed explanations of each action's execution. Similarly, we perform 20 simulations for each action taken with \name, and increase this number to 50 for thought revision processes.

In the multi-solution scenario, all of the IO, CoT, and CoT-SC prompts consist of four examples. Each problem is presented with one to three distinct solutions. For ToT (b=3) and GoT (k=3), the maximum number of steps is increased to 12, as correct solutions may not always be optimal and could exceed 9 steps. In the case of \name, after conducting simulations with MCTS, we sample 50 thought trajectories for exploration and select the top-3 thoughts with the highest counts.

\subsubsection{Results}
\begin{table}[t]
\centering
\small
\caption{Performance comparison on 8-Puzzle. 
\label{tab:8puzzle.with_laststep}}
\resizebox{0.9\width}{!}{
\begin{tabular}{l|ccc|ccc} 
\hline
\multirow{2}{*}{\textbf{Model}} & \multicolumn{3}{c|}{\textbf{GPT-3.5}}                          & \multicolumn{3}{c}{\textbf{GPT-4}}                              \\ 
\cline{2-7}
& \multicolumn{1}{p{1.2cm}}{\centering\scriptsize\textbf{Acc. [\%]}} & \multicolumn{1}{p{1.2cm}}{\centering\scriptsize\textbf{LLM invoked}} & \multicolumn{1}{p{1.2cm}|}{\centering\scriptsize\textbf{$f_\theta$ \\invoked}} & \multicolumn{1}{p{1.2cm}}{\centering\scriptsize\textbf{Acc. [\%]}} & \multicolumn{1}{p{1.2cm}}{\centering\scriptsize\textbf{LLM invoked}} & \multicolumn{1}{p{1.2cm}}{\centering\scriptsize\textbf{$f_\theta$ \\invoked}}   \\ 
\hline
IO                              & 0.00               & 1.00                 & -                           & 1.68               & 1.00                 & -                            \\
CoT                             & 0.00               & 1.00                 & -                           & 7.56               & 1.00                 & -                            \\
CoT-SC                  & 0.84               & 10.00                & -                           & 8.40               & 10.00                & -                            \\
ToT (b=1)                       & 5.88               & 31.76                & -                           & 3.36               & 27.49                & -                            \\
ToT (b=3)                       & 6.72               & 55.86                & -                           & 13.45              & 54.13                & -                            \\
GoT (k=1)                       & 3.36               & 19.00                & -                           & 3.36               & 19.00                & -                            \\ 
LLaMA-2-13B                        & 0.00               & -                 & -                  & 0.00              & -                 & -                   \\ 
MCTS                        & 51.26               & -                 & -                  & 51.26             & -                 & -                   \\ 
\hline
\textbf{XoT (w/ 1 r)}        & \textbf{59.66}     & \textbf{1.50}        & \textbf{41.09}              & \textbf{93.28}     & \textbf{1.48}        & \textbf{55.66}               \\
\textbf{XoT (w/ 2 r)}        & \textbf{59.66}     & \textbf{1.92}        & \textbf{42.18}              & \textbf{94.96}     & \textbf{1.55}        & \textbf{58.91}               \\
\textbf{XoT (w/ 3 r)}        & \textbf{63.03}     & \textbf{2.29}        & \textbf{42.60}              & \textbf{95.80}     & \textbf{1.61}        & \textbf{62.22}               \\
\hline
\end{tabular}}
\end{table}

The inherent spatial complexity of the 8-Puzzle, the need for long-term planning, and the presence of invalid actions create a significant challenge for LLMs, which rely solely on textual data as input. This challenge is starkly evident in the poor performance of the baselines on both GPT-3.5, where its IO prompting achieve a mere 0\% success rate. \name successfully addresses this issue by supplying thoughts acquired from MCTS, thereby infusing external knowledge into the problem-solving process. This augmentation empowers LLMs to tackle problems that were previously insurmountable. In summary, when using GPT-4, \name achieves an accuracy of 93.28\% with 1 revision and 95.80\% with 3 revisions in the 8-Puzzle task, outperforming the best prompting baseline, ToT (b=3), which only achieves 13.45\% accuracy. Additionally, \name demonstrates efficiency, as it only requires approximately 1.6 LLM calls for 3-time revision setting. The poor performance of finetuned LLaMA-2-13B (0\%) revealed a significant issue with hallucination. This  underscores the inefficiency and ineffectiveness of finetuning approaches for tasks necessitating long-term planning, while also bringing to light the heightened costs associated with its use.

The multi-solution performance presented in Table~\ref{tab:8puzzle.multi} confirms that the \name method continues to outperform other baselines for both GPT-3.5 and GPT-4 models in terms of  MultiAcc, whether or not revision is applied. The revision process of \name is particularly beneficial for GPT-4, as it improves the MultiAcc from 51.26\% to 76.33\%, compared to single MCTS. These results again demonstrate that \name can effectively generate complex thought structures for multi-solutions with high performance and efficiency, making it particularly suitable for this task.

\begin{table}[t]
\centering
\small
\caption{Performance comparison on 8-Puzzle in the multi-solution scenario.
\label{tab:8puzzle.multi}}
\resizebox{0.9\textwidth}{!}{
\begin{tabular}{l|cccc|cccc} 
\hline
\multirow{2}{*}{\textbf{Model}} & \multicolumn{4}{c|}{\textbf{GPT-3.5}}                                                                                                      & \multicolumn{4}{c}{\textbf{GPT-4}}                                                                                                          \\ 
\cline{2-9}
                                & \begin{tabular}[c]{@{}c@{}}\textbf{Multi }\\\textbf{Acc.}\end{tabular}  & \textbf{\#Sol} & \begin{tabular}[c]{@{}c@{}}\textbf{LLM }\\\textbf{invoked}\end{tabular} & \begin{tabular}[c]{@{}c@{}}\textbf{$f_\theta$}\\\textbf{invoked}\end{tabular}  & \begin{tabular}[c]{@{}c@{}}\textbf{Multi }\\\textbf{Acc.}\end{tabular} & \textbf{\#Sol} & \begin{tabular}[c]{@{}c@{}}\textbf{LLM }\\\textbf{invoked}\end{tabular} & \begin{tabular}[c]{@{}c@{}}\textbf{$f_\theta$}\\\textbf{invoked}\end{tabular}  \\ 
\hline
IO                              & 0.00              & 2.47           & 1.00                                                                    & -                           & 0.84              & 2.97           & 1.00                                                                    & -                            \\
CoT                             & 1.43              & 2.05           & 1.00                                                                    & -                           & 7.84              & 1.21           & 1.00                                                                    & -                            \\
CoT-SC               & 1.54              & 1.90           & 10.00                                                                   & -                           & 6.58              & 2.08           & 10.00                                                                   & -                            \\
ToT (b=3)                       & 2.52              & 2.98           & 55.86                                                                   & -                           & 5.60              & 2.97           & 54.13                                                                   & -                            \\
GoT (k=3)                       & 3.36              & 2.96           & 24.18                                                                   & -                           & 16.61             & 2.70           & 22.76                                                                   & -                            \\ 
\hline
\textbf{XoT (w/ 1 r)}        & \textbf{27.45}    & \textbf{2.85}  & \textbf{4.19}                                                           & \textbf{52.06}              & \textbf{76.33}    & \textbf{1.52}  & \textbf{4.30}                                                           & \textbf{66.66}               \\
\hline
\end{tabular}}
\end{table}

\begin{table}[t]
\centering
\small
\caption{Performance comparison on Pocket Cube. \label{tab:cube.with_laststep}}
\resizebox{0.9\width}{!}{
\begin{tabular}{l|ccc|ccc} 
\hline
\multirow{2}{*}{\textbf{Model}} & \multicolumn{3}{c|}{\textbf{GPT-3.5}}                          & \multicolumn{3}{c}{\textbf{GPT-4}}                              \\ 
\cline{2-7}
& \multicolumn{1}{p{1.2cm}}{\centering\scriptsize\textbf{Acc. [\%]}} & \multicolumn{1}{p{1.2cm}}{\centering\scriptsize\textbf{LLM invoked}} & \multicolumn{1}{p{1.2cm}|}{\centering\scriptsize\textbf{$f_\theta$ \\invoked}} & \multicolumn{1}{p{1.2cm}}{\centering\scriptsize\textbf{Acc. [\%]}} & \multicolumn{1}{p{1.2cm}}{\centering\scriptsize\textbf{LLM invoked}} & \multicolumn{1}{p{1.2cm}}{\centering\scriptsize\textbf{$f_\theta$ \\invoked}}   \\ 
\hline
IO                              & 1.09               & 1.00                 & -                           & 1.09               & 1.00                 & -                            \\
CoT                             & 0.00               & 1.00                 & -                           & 1.09               & 1.00                 & -                            \\
CoT-SC                  & 0.00               & 10.00                & -                           & 1.09               & 10.00                & -                            \\
ToT (b=1)                       & 7.65               & 16.50                & -                           & 11.48              & 16.39                & -                            \\
ToT (b=3)                       & 17.49              & 58.72                & -                           & 19.57              & 56.58                & -                            \\
GoT (k=1)                       & 1.64               & 8.93                 & -                           & 18.03              & 8.55                 & -                            \\ 
LLaMA-2-13B                        & 0.00               & -                 & -                  & 0.00              & -                 & -                   \\ 
MCTS                        & 46.44               & -                 & -                  & 46.44             & -                 & -                   \\ 
\hline
\textbf{XoT (w/ 1 r)}        & \textbf{74.32}     & \textbf{1.55}        & \textbf{64.63}              & \textbf{77.60}     & \textbf{1.54}        & \textbf{75.51}               \\
\textbf{XoT (w/ 2 r)}        & \textbf{80.33}     & \textbf{1.81}        & \textbf{96.46}              & \textbf{79.32}     & \textbf{1.79}        & \textbf{146.52}               \\
\textbf{XoT (w/ 3 r)}        & \textbf{84.70}     & \textbf{2.01}        & \textbf{103.22}              & \textbf{83.61}     & \textbf{2.00}        & \textbf{84.63}               \\
\hline
\end{tabular}
}
\end{table}

\begin{table}[t]
\centering
\small
\caption{Performance comparison on Pocket Cube in the multi-solution scenario. \label{tab:cube.multi}}
\resizebox{0.9\textwidth}{!}{
\begin{tabular}{l|cccc|cccc} 
\hline
\multirow{2}{*}{\textbf{Model}} & \multicolumn{4}{c|}{\textbf{GPT-3.5}}                                                                                                      & \multicolumn{4}{c}{\textbf{GPT-4}}                                                                                                          \\ 
\cline{2-9}
                                & \begin{tabular}[c]{@{}c@{}}\textbf{Multi }\\\textbf{Acc.}\end{tabular}  & \textbf{\#Sol} & \begin{tabular}[c]{@{}c@{}}\textbf{LLM }\\\textbf{invoked}\end{tabular} & \begin{tabular}[c]{@{}c@{}}\textbf{$f_\theta$}\\\textbf{invoked}\end{tabular}  & \begin{tabular}[c]{@{}c@{}}\textbf{Multi }\\\textbf{Acc.}\end{tabular} & \textbf{\#Sol} & \begin{tabular}[c]{@{}c@{}}\textbf{LLM }\\\textbf{invoked}\end{tabular} & \begin{tabular}[c]{@{}c@{}}\textbf{$f_\theta$}\\\textbf{invoked}\end{tabular}  \\ 
\hline
IO                              & 0.27              & 2.00           & 1.00                                                                    & -                           & 1.09              & 1.98           & 1.00                                                                    & -                            \\
CoT                             & 0.55              & 1.05           & 1.00                                                                    & -                           & 0.82              & 1.91           & 1.00                                                                    & -                            \\
CoT-SC                  & 0.18              & 2.90           & 10.00                                                                   & -                           & 0.82              & 2.92           & 1.00                                                                    & -                            \\
ToT (b=3)                       & 5.83              & 2.99           & 58.72                                                                   & -                           & 6.52              & 2.99           & 56.58                                                                   & -                            \\
GoT (k=3)                       & 1.09              & 2.99           & 14.76                                                                   & -                           & 16.85             & 2.77           & 13.36                                                                   & -                            \\ 
\hline
\textbf{XoT (w/ 1 r)}        & \textbf{48.72}    & \textbf{2.20}  & \textbf{4.13}                                                           & \textbf{115.73}             & \textbf{77.41}    & \textbf{1.72}  & \textbf{4.08}                                                           & \textbf{122.54}              \\
\hline
\end{tabular}}
\end{table}

\subsection{Pocket Cube}
The $2\times 2$ Pocket Cube is a simplified variant of the classic Rubik's Cube puzzle. Its primary objective is to restore all of its faces to a uniform color by executing various face rotations. The maximum number of steps required to optimally solve the cube is 11, and it is also a NP-complete problem \cite{demaine2017solving}  and may possess multiple solutions. This task is known to be challenging to LLMs \cite{cube}.

\subsubsection{Task Setup}
We initially set all faces of the cube to a uniform color and then randomly apply 5 actions sequentially selected from the 27 legal actions of the Rubik's Cube. This process resulted in the creation of 1,000 training samples and 183 testing samples. All generated problems can be solved within 4 steps. To simplify the action space, we reduced the 27 legal operations to 9 actions, namely: \{U, U', U2, R, R', R2, F, F', F2\}, which are used in our experiments with both baselines and \name. As shown in Table~\ref{tab:compare}, the thoughts pertain to the step-by-step rotation, and the cube state after the move.


\subsubsection{Baselines \& \name Setup} 
The IO prompt is augmented with a single in-context example. In CoT, we enrich each input-output pair by including intermediate actions and states. In ToT, we retrieve one-step thought candidates from the LLM at each stage and instruct the LLM to classify each candidate for intermediate selection. 
A maximum step limit of 4 is imposed, as all generated problems can be resolved within this range. 
The cube's rules are conveyed through a system message, which includes the definition of the action space and illustrations of the execution of each action. For \name, we conduct 20 simulations for each action taken and increase it to 500 for revision.

In the multi-solution setup, the IO, CoT, and CoT-SC prompts each include 3 examples, and each problem within these prompts offers 3 unique solutions. As for ToT (b=3) and GoT (k=3), the maximum number of steps allowed is extended to 7. In the case of \name, after conducting MCTS simulations, we gather 50 thought trajectories, and we keep the top 3 thoughts with the highest counts.

\subsubsection{Results}

The Pocket Cube task, similar to the 8-Puzzle, poses a challenge that demands spatial imagination skills, making it difficult for LLMs to excel. As expected, most of the baselines show very poor performance in this task, with some baselines achieving 0\% accuracy. The best prompting baseline, ToT (b=3) with GPT-4, only attains a success rate of 19.57\%. In contrast, \name can achieve over 77.60\% accuracy with 1-time revision and over 80\% accuracy with 3-time revision, establishing itself as an expert in solving this task. This is attributed to the injection of external knowledge from MCTS, enabling LLMs to solve problems that they would struggle with on their own. On the other hand, \name improves accuracy by 30\% compared to a single MCTS with one-time revision. This demonstrates the effectiveness of integrating MCTS and LLMs. Notably, \name maintains high efficiency in this task, requiring only approximately 2 LLM inference calls for both GPT-3.5 and GPT-4. 
Again, the finetuned LLaMA-2-13B struggles with the Pocket Cube task (0\%), due to significant hallucination issues. This comparison further validates the potential of \name in contexts demanding extensive planning and decision-making accuracy.

In the case of the multi-solution scenario, the performance of the \name method remains remarkable, achieving over 77\% MultiAcc with GPT-4. The revision process continues to play an important role, significantly improving the performance of \name with both GPT models. The closest competitor in this setting is GoT (k=3) with GPT-4, which achieves a MultiAcc of 16.85\%, but it requires a significantly higher number of LLM invocations compared to \name (13.36 vs. 4.08) and much lower MultiAcc. Overall, \name retains its position as the best solution for the Pocket Cube.

\subsection{Ablation Study\label{sec.ablation}}
In our ablation study, we consider two aspects: the impact of the number of revisions on the performance and efficiency of \name and the sensitivity of performance to the completeness of the provided thoughts. These angles allow us to gain insights into how \name's performance can be improved and understand the importance of providing complete thoughts in complex problem-solving tasks.

\subsubsection{Number of Revisions}
\begin{figure}[t]
\centering
\subfigure[Game of 24]{
\includegraphics[width=0.9\columnwidth]{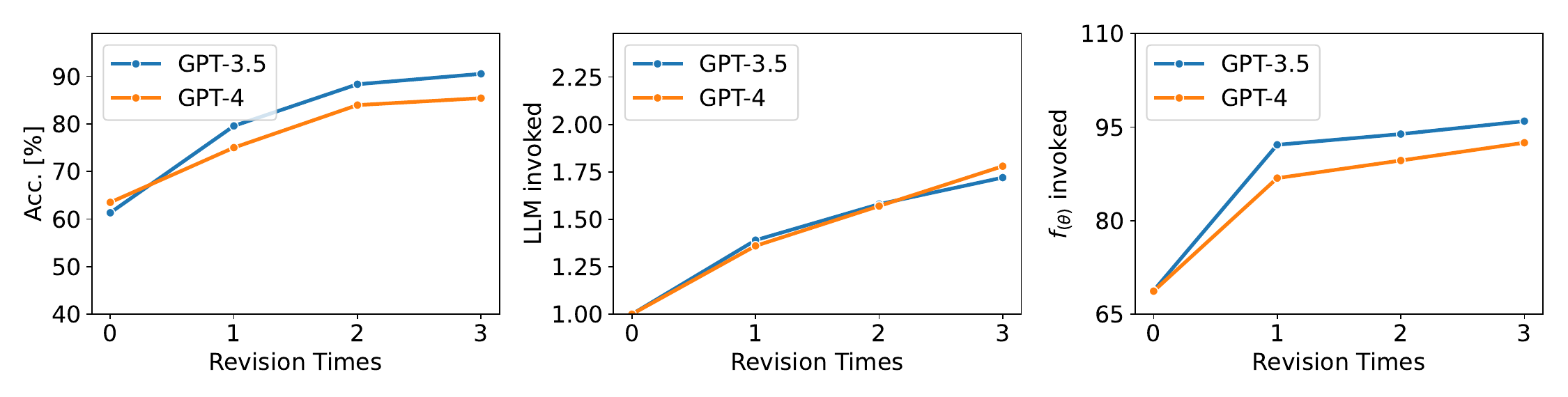}}
\subfigure[8-Puzzle]{
\includegraphics[width=0.9\columnwidth]{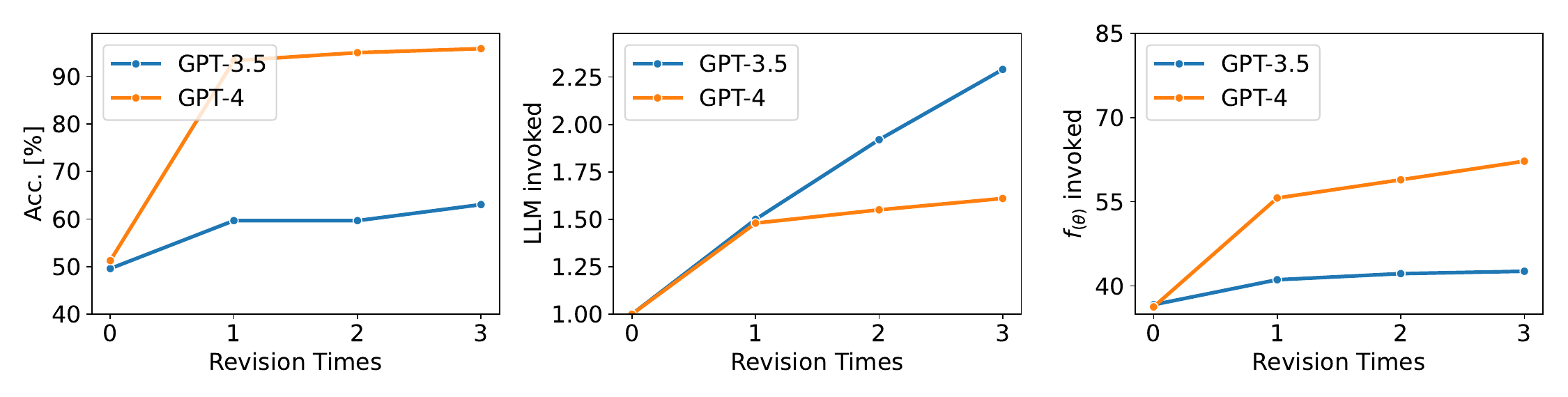}}

\subfigure[Pocket Cube]{
\includegraphics[width=0.9\columnwidth]{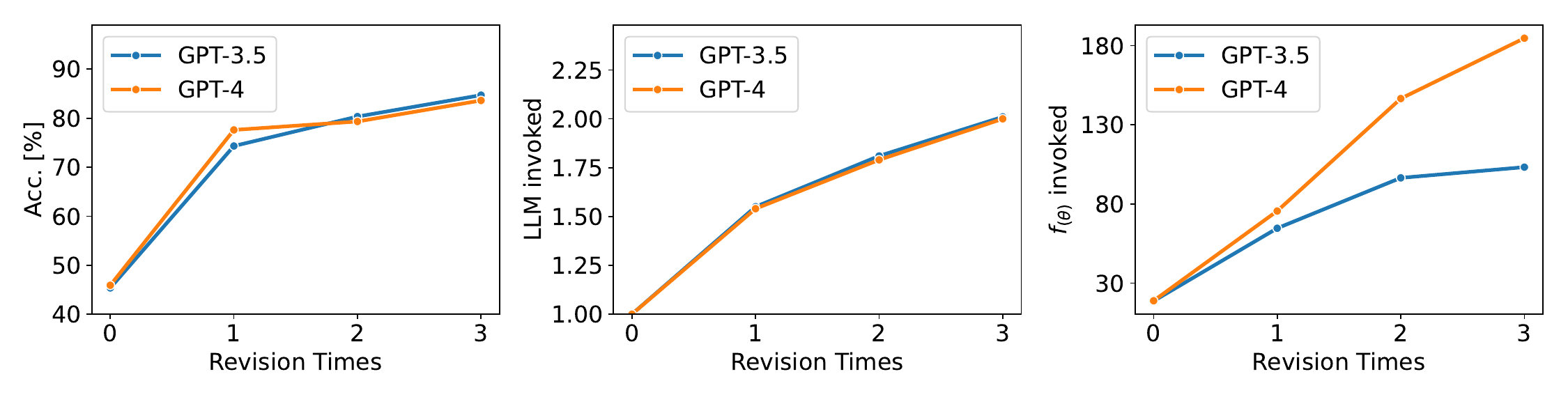}}

\caption{Accuracy, LLM and $f_\theta$ invoked comparison on \name w.r.t. the number of revisions.
\label{fig:ablation}}
\end{figure}

It's important to highlight that the performance of each task can be further improved through multiple revisions of the thought using the MCTS-LLM collaborative framework. In Fig.~\ref{fig:ablation}, we compare the performance of GPT-3.5 and GPT-4 models using the \name method with varying numbers of revisions, ranging from 0 to 3, across all three tasks.

In the Game of 24 task, as the number of revisions increases, both models exhibit improved performance. Notably, GPT-3.5 consistently outperforms GPT-4 in terms of accuracy. After three revisions, GPT-3.5 achieves an accuracy of 90.51\%, while GPT-4 reaches 85.40\%. This improved performance comes at the cost of increased inference times and model calls, primarily driven by the need for more interactions to generate revised thoughts.
For the 8-Puzzle task, the trend of increasing accuracy with more revisions remains valid. However, in this task, GPT-4 significantly outperforms GPT-3.5. After one revision, GPT-4 achieves an accuracy of 93.28\%, which increases to 95.80\% after the third revision. In contrast, GPT-3.5 only attains an accuracy of 63.03\% after the third revision.
In the Pocket Cube task, the performance trend is similar. The accuracy of both models improves with an increase in the number of revisions. GPT-3.5 starts at an accuracy of 45.36\% without revision and improves to 84.70\% after three revisions. GPT-4 begins with an accuracy of 45.90\% and reaches 83.61\% after three revisions. Inference times and model calls are comparable between the two models, with GPT-4 showing a substantial increase in model calls after the third revision.

Note that the number of LLM invocations does not increase dramatically with additional revisions, even though $f_\theta$ is called more times to guide simulations. Considering the significant disparity in inference costs between LLM and $f_\theta$, increasing the number of revisions to achieve better performance appears to be a favorable trade-off.

\begin{table}[!t]
\small
\centering
\caption{Revision Success Rate for GPT-3.5.}
\label{tab:error_detection_gpt35}
\begin{tabular}{lccc}
\hline
\textbf{Revisions} & \textbf{Game of 24} & \textbf{8-Puzzle} & \textbf{Pocket Cube} \\ \hline
XoT (w/ 1 r) & 47.17\% & 20.00\% & 53.00\% \\
XoT (w/ 2 r) & 69.81\% & 21.31\% & 63.64\% \\
XoT (w/ 3 r) & 75.93\% & 26.67\% & 72.00\% \\ \hline
\end{tabular}
\end{table}

\begin{table}[!t]
\small
\centering
\caption{Revision Success Rate for GPT-4.}
\label{tab:error_detection_gpt4}
\begin{tabular}{lccc}
\hline
\textbf{Revisions} & \textbf{Game of 24} & \textbf{8-Puzzle} & \textbf{Pocket Cube} \\ \hline
XoT (w/ 1 r) & 32.69\% & 85.96\% & 58.59\% \\
XoT (w/ 2 r) & 55.10\% & 89.47\% & 60.00\% \\
XoT (w/ 3 r) & 60.00\% & 91.38\% & 70.00\% \\ \hline
\end{tabular}
\end{table}

We also focus on the efficacy of the revision process within the \name framework across three distinct tasks. The Revision Success Rate is calculated as the ratio of successfully detected errors to the number of failed cases without revision, thereby providing insight into the effectiveness of revisions. The results for both GPT-3.5 and GPT-4 are presented in Table~\ref{tab:error_detection_gpt35} and Table~\ref{tab:error_detection_gpt4}. Our observations reveal a high revision success rate in the XoT framework, which increases with the number of revisions. This underscores the effectiveness of LLMs in the revision process, positioning it as a highly efficient approach to thoughts revision.

\begin{table}[htbp]
\centering
\small
\caption{Performance comparison on three tasks with incomplete thoughts. \label{tab:w/olaststep}}
\resizebox{1\textwidth}{!}{
\begin{tabular}{l|l|ccc|ccc} 
\hline
\multirow{2}{*}{\textbf{ Task}}       & \multirow{2}{*}{\textbf{ Model}} & \multicolumn{3}{c|}{\textbf{GPT-3.5}}                                   & \multicolumn{3}{c}{\textbf{GPT-4}}                                       \\ 
\cline{3-8}
                                      &                                  & \textbf{Acc. [\%]} & \textbf{LLM invoked} & \textbf{$f_\theta$ invoked} & \textbf{Acc. [\%]} & \textbf{LLM invoked} & \textbf{$f_\theta$ invoked}  \\ 
\hline
\multirow{3}{*}{Game of 24}  & ToT (b=1)                        & 3.65               & 17.15                & -                           & 40.88              & 18.55                & -                            \\
                                      & GoT (k=1)                        & 2.19               & 5.00                 & -                           & 9.49               & 5.00                 & -                            \\
                                      & \textbf{XoT (w/o revise)}        & \textbf{17.52}     & \textbf{1.00}        & \textbf{68.73}              & \textbf{43.07}     & \textbf{1.00}        & \textbf{68.70}               \\ 
\hline
\multirow{3}{*}{8-Puzzle}    & ToT (b=1)                        & 0.00               & 32.60                & -                           & 6.72               & 26.98                & -                            \\
                                      & GoT (k=1)                        & 0.00               & 18.63                & -                           & 3.36               & 19.00                & -                            \\
                                      & \textbf{XoT (w/o revise)}        & \textbf{2.52}      & \textbf{1.00}        & \textbf{36.66}              & \textbf{40.34}     & \textbf{1.00}        & \textbf{36.24}               \\ 
\hline
\multirow{3}{*}{Pocket Cube} & ToT (b=1)                        & 0.55               & 16.48                & -                           & 2.19               & 16.39                & -                            \\
                                      & GoT (k=1)                        & 0.00               & 8.96                 & -                           & 1.64               & 8.68                 & -                            \\
                                      & \textbf{XoT (w/o revise)}        & \textbf{5.46}      & \textbf{1.00}        & \textbf{18.85}              & \textbf{6.01}      & \textbf{1.00}        & \textbf{18.89}               \\
\hline
\end{tabular}}
\end{table}

\subsubsection{Incomplete Thought}
In this ablation study, we explore the performance of LLMs when provided with incomplete thoughts, specifically omitting the last step of the thought trajectory. This simulates scenarios where MCTS might supply inaccurate or incomplete thoughts. The aim is to test whether LLMs can independently solve problems or rely on their own reasoning, rather than solely relying on the thought from MCTS as answers. We present the performance comparison for all three tasks in Table~\ref{tab:w/olaststep}. Note that we only compare ToT and GoT since other baselines do not support this comparison by their nature. 

The results clearly show that incomplete thoughts lead to a significant performance drop in all three tasks. GPT-3.5 is more affected than GPT-4, with GPT-3.5 achieving 0\% accuracy on several baselines. In contrast, \name with GPT-4 attains satisfactory performance on the Game of 24 and 8-Puzzle, achieving over 40\% accuracy. However, the performance of \name is dramatically affected in the Pocket Cube task, with accuracy dropping to 6\%. This demonstrates that for very complex tasks, LLMs are highly sensitive to the completeness of the thoughts provided. Missing steps in the thought can lead to a substantial drop in performance, highlighting the importance of providing complete thoughts for such tasks.

\subsection{Case Study\label{sec.case_study}}
\begin{figure}[t]
\centering
\includegraphics[width=1\columnwidth]{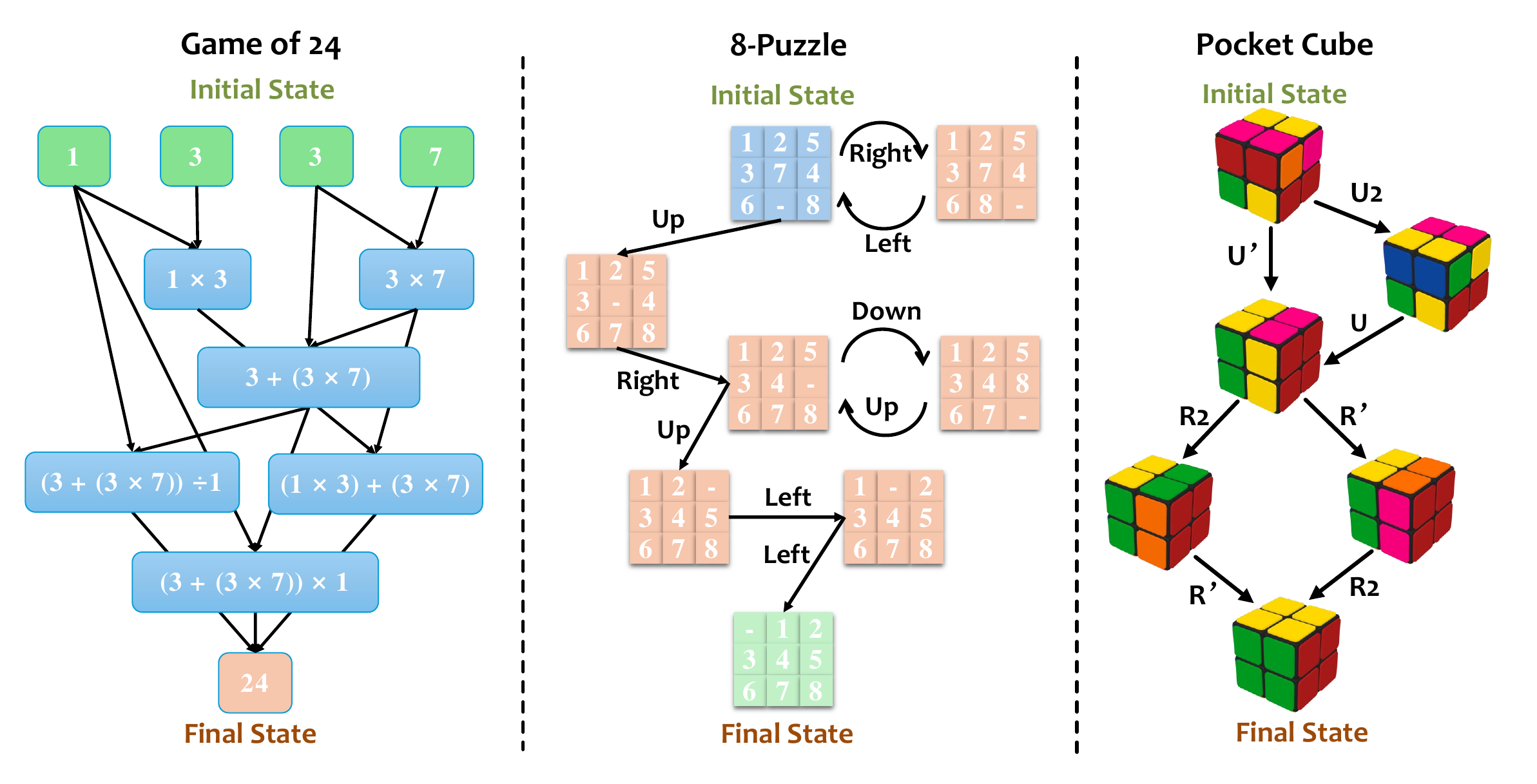}
\vspace*{-2.5em}
\caption{Examples of thought structures generated by \name for all three tasks in the multi-solution scenario.
\label{fig:case}}

\end{figure}

Finally, in Fig.~\ref{fig:case}, we provide examples of thought structures generated by \name for all three tasks in the multi-solution scenario. It is noteworthy that, owing to the multiple solutions required, the generated thoughts intertwine during intermediate steps and converge towards the final goal state. This results in a naturally woven thought structure resembling a graph, showcasing the remarkable flexibility achieved by \name. Upon closer examination of each example, in the case of the Game of 24, there are multiple solutions to reach the goal of 24 from the initial state. \name effectively predicts these trajectories, indicating its ability to grasp complex thought structures. In the 8-Puzzle example, we observe instances of reflection in the thought structure, with back-and-forth recurrent state transitions. This demonstrates \name's capacity for self-reflection, a crucial attribute for LLMs, as discussed in previous work \cite{shinn2023reflexion}. In the case of the Pocket Cube, \name identifies four distinct pathways to reach the goal state, leading to successful problem-solving across multiple solutions.

Overall, these cases highlight how \name encapsulates the flexibility required in thought generation, fostering diverse and creative thinking for LLMs. This enables them to produce multiple high-quality answers to a single problem effectively.

\subsection{Experiment Summary}
In summary, our approach \name significantly improves the performance of LLMs by introducing a streamlined thought trajectory revision process. This represents a fundamental shift from traditional problem-solving approaches, resulting in substantial performance enhancements across a range of tasks. Notably, \name excels in solving the Game of 24 and demonstrates its ability to overcome challenges requiring spatial reasoning, such as the 8-Puzzle and Pocket Cube, which were previously challenging for LLMs. The remarkable synergy of improved performance, efficiency, and flexibility exhibited by \name positions it as an exemplary and superior method for eliciting optimal responses from LLMs.

\section{Related Work}
\textbf{Decision Making \& Planning with LLMs.}
The utilization of LLMs for decision-making and planning has become a prominent area of research. Similar to human problem-solving, the process involves breaking down complex problems into sub-tasks. Various frameworks, such as CoT \cite{wei2022chain}, ToT \cite{yao2023tree}, and GoT \cite{besta2023graph}, have been designed to facilitate problem decomposition in different structural forms, leading to enhanced solutions derived from LLMs. Extensions of these frameworks have also been explored across different domains and modalities \cite{zhang2022automatic, zhang2023multimodal, ning2023skeleton, turpin2023language, long2023large}. Our approach \name distinguishes itself from the aforementioned work by concurrently achieving superior performance, efficiency, and flexibility, embodying the concept of comprehensive thought generation.

Furthermore, the ``Describe, Explain, Plan, and Select'' framework introduced in \cite{wang2023describe} presents an interactive planning approach for LLMs, significantly enhancing planning performance for multi-task agents. Research conducted in \cite{singh2023progprompt} leverages LLMs to suggest next actions or sequences during task planning for robotics, leading to improved task performance across various metrics. Additionally, work presented in \cite{xie2023translating} employs LLMs to translate natural language into planning goals, demonstrating their capacity to harness commonsense knowledge and reasoning to provide missing details for under-specified goals. These studies underscore the growing potential of LLMs in the field of planning, with research efforts expanding rapidly.

\textbf{Augmenting LLMs with RL.}
Enhancing the capabilities of LLMs through the incorporation of external models constitutes an effective strategy for improving their overall quality. The foundational work of ChatGPT \cite{ouyang2022training} leverages RL from human feedback to enable LLMs to adhere to human guidance, resulting in a substantial enhancement of their truthfulness and a reduction in toxic output. Similarly, GLAM \cite{carta2023grounding} employs online RL to establish alignment between LLMs' knowledge and the broader environment, thus enhancing their ability to generalize to new objects or tasks and ultimately improving their performance. Additionally, an interesting study in \cite{yuan2023plan4mc} utilizes RL to acquire basic skills in the context of Minecraft \cite{cipollone2014minecraft}, with subsequent high-level planning carried out by LLMs. This approach demonstrates promising performance across various Minecraft tasks. Furthermore, the ESPER framework \cite{yu2023fusing} harnesses RL to achieve alignment between multimodal inputs and language model generations, all without the need for direct supervision. This empowers LLMs to effectively tackle multimodal tasks and provides robust visual alignment and rapid inference speeds while preserving the textual domain. Collectively, these research endeavors underscore the considerable potential in augmenting LLMs with reinforcement learning techniques.

MCTS is also integrated with LLMs to enhance both training and inference processes. Hao \etal propose ``Reasoning via Planning'', utilizing LLMs as a world model and reasoning agent, while combining MCTS as a strategic explorer to enhance LLMs' reasoning and planning abilities \cite{hao2023reasoning}. Liu \etal incorporate MCTS and PPO \cite{schulman2017proximal} to devise a value-guided decoding algorithm, thereby enhancing the preferability of generated text by LLMs \cite{liu2023making}. Additionally, Feng \etal employ MCTS to augment LLMs' decoding and, consequently, their reasoning and planning capabilities \cite{feng2023alphazero}. These studies underscore the significant potential of integrating MCTS with LLMs to improve their overall capabilities.

\section{Discussion}

\noindent \textbf{Generalization} 
While \name is presently utilized for reasoning and search problems, its applicability can be extended to a broader spectrum of problem domains characterized by decomposable tasks with well-defined objectives. The MCTS utilized in \name is particularly suitable for such tasks and can therefore generalize to more complex problems. We also note that MCTS is functioning in a supportive role and can be substituted with alternative supervised or RL models for thought exploration and generation, which can serve as a copilot to inject domain knowledge of the real-world model to LLMs. This opens up a promising avenue for future research, enabling LLMs to engage in more effective planning and problem solving processes.

\noindent \textbf{Limitation} 
We also note that the implementation of \name necessitates the training of additional policy and value models to expedite the inference process. This training process requires the acquisition of datasets from real-world environments, introducing supplementary costs and efforts. However, note that these policy and value models are considerably smaller and more computationally efficient than the underlying LLMs. Consequently, the incurred costs are deemed low, particularly in the context of tasks featured in this study, where the thought steps and objectives are well-defined. In future research endeavors, we intend to explore methods to enhance the efficiency of the training process for \name in scenarios where the objectives are less straightforward, such as multi-agent planning and code generation tasks \cite{talebirad2023multi, vaithilingam2022expectation}. This endeavor will expand the applicability of the proposed \name framework to a broader range of applications.

In terms of potential risks, \name is susceptible to the MCTS module providing incorrect intermediate thoughts, which may result in an inaccurate final answer or hallucination. Changes in the environment could lead to inaccuracies in MCTS and subsequently in the thoughts provided to LLMs. However, LLMs have proven effective in revising thoughts by leveraging their internal knowledge, mitigating the risk associated with inaccuracies in the initial thought generation. Additionally, LLMs may make mistakes and sometimes deviate from the thoughts generated by the MCTS module, leading to errors. This aspect should be taken into consideration when employing the approach.

\noindent \textbf{Conclusion}
The \name framework presented in this paper signifies a significant progression in thought generation for LLMs aimed at solving complex tasks. It challenges the constraints of the ``Penrose Triangle \Itriangle'' by concurrently achieving performance, efficiency, and flexibility, a feat unattainable by existing prompting paradigms. This accomplishment is achieved through the integration of MCTS with pretrained low-cost policy and value networks, by injecting domain knowledge  and planning capability into LLMs, offloading thought searching, and facilitating unconstrained free-style thought exploration. The collaborative thought revision framework involving MCTS and LLM further enhances the quality of thought generation.
Experimental evaluations conducted across three intricate real-world problems, namely the Game of 24, 8-Puzzle, and Pocket Cube, provide empirical evidence that our \name framework significantly outperforms existing prompting paradigms, particularly in scenarios involving multi-solution problems.

\bibliography{iclr2024_conference}
\bibliographystyle{iclr2024_conference}

\appendix

\section{LLaMA-2-13B Setup\label{appendix.baseline}} 

\noindent \textbf{LLaMA-2-13B (finetuned).} To evaluate the potential of directly distilling knowledge from simulations into a smaller model to possibly avoid using a large model like GPT-4 during testing, we fine-tuned the LLaMA-2-13B model. Our experiments were carried out on eight V100 GPUs, each with 80GB of memory, and lasted approximately 5 hours. The training setup involved 5 epochs, a train batch size of 32, an evaluation batch size of 1, and a single step for gradient accumulation. The evaluation and save strategies were set to "no" and "steps" respectively, with saving occurring every 20 steps and a limit of one saved model. The learning rate was 2e-5, with no warmup steps and logging every 2 steps. We employed a cosine learning rate scheduler. By using ground truth labels—considered more accurate than labels from MCTS simulations—we aimed to convert an optimization or search problem into a more straightforward prediction or supervised learning challenge, using a training dataset of (question, answer) pairs.

\section{Computational Training Costs of MCTS \label{appendix.training_cost}}

The number of training and testing policy/value model calls for XoT are listed in Table~\ref{tab:model_calls}. We train this model through three iterations, each comprising 10 self-play episodes for MCTS. Offline pretraining serves as a one-time solution that reduces the computational burden of testing by integrating external knowledge. Methods like ToT and GoT, which rely solely on the LLMs' internal knowledge, do not require pretraining but necessitate frequent calls to LLM during testing. For example, the average number of LLM invocations for three tasks in ToT are 39.83, 54.13, and 56.58, averaging 50.18 times per test problem. The computational cost of these recurring calls during testing exceeds the pretraining cost of the policy/value model in XoT.

Futhermore, it's worth highlighting that GPT-3.5 boasts 175 billion parameters, and GPT-4 is estimated to have an astonishing over 1 trillion parameters. In contrast, the total number of parameters in the Policy/Value Network for all three tasks is approximately 1e6. This deliberate design choice results in a model significantly smaller than LLMs, ensuring efficiency even with additional calls during training.

\section{Experiment Results on other NLP tasks\label{appendix.nlp}}

In addition to the tasks employed in this paper, many other NLP tasks can be formulated as MCTS searching problems, using LLMs to get rewards and rendering XoT applicable to a broader range of scenarios. For example, in ToT \cite{yao2023tree}, the task of Creative Writing uses LLMs to evaluate the quality of generated paragraphs. In a similar vein, GoT \cite{besta2023graph} utilizes LLMs to rate the outcomes of Document Merging tasks. This strategy of employing LLMs for reward design is gaining traction and is currently a subject of active research \cite{kwon2023reward}.

\begin{table}[t]

    \centering
    \caption{Number of policy/value model calls in training and testing per iteration for different tasks.\label{tab:model_calls}}
    \begin{tabular}{lccc}
        \toprule
        & \textbf{Game of 24} & \textbf{8-Puzzle} & \textbf{Pocket Cube} \\
        \midrule
        Training & 1044.70 & 834.70 & 787.00 \\
        Testing & 88.20 & 55.66 & 75.51 \\
        \bottomrule
    \end{tabular} 
\end{table}

\begin{table}[t]
\caption{Performance comparison on Document Merging.\label{tab:method_comparison}}
\centering
\begin{tabular}{lcc}
\hline
\textbf{Method} & \textbf{Score (0-10)} & \textbf{Cost (Avg num of tokens)} \\ \hline
IO              & 6.390                 & 2292.60                           \\
CoT             & 6.524                 & 3152.90                           \\
ToT & 7.715                 & 51486.00                          \\
GoT             & 7.559                 & 27685.28                          \\
XoT             & 8.168                 & 15270.80                          \\ \hline
\end{tabular}
\end{table}

To illustrate, we present preliminary results for GPT-3.5 on the Document Merging task in Table \ref{tab:method_comparison}, where the scores are indicative of a weighted combination of duplication and information intact in the merged document (the higher the better). The objective of this task is to create a new Non-Disclosure Agreement (NDA) document by combining several input documents that partially overlap in content. The aim is to minimize duplication while maximizing information retention. The experimental setting is aligned with in the GoT \cite{besta2023graph} paper. We utilized the same dataset provided in their repository.

Remarkably, XoT emerges as the most effective approach, achieving the highest score of 8.168. Notably, XoT maintains a balance in resource efficiency, with an average token cost of 15270.80, surpassing both ToT and GoT. These outcomes underscore XoT's advanced capabilities in handling general textual tasks, extending beyond gaming problems.

\section{Prompt Example}
Prompts 1-3 display example CoT prompts utilized for Game of 24, 8-Puzzle, and Pocket Cube. These templates are applicable to CoT, ToT, GoT, and our \name in the final inference process. Each thought step includes the action taken and the resulting new state.

\begin{tcolorbox}[colback=black!5!white,colframe=black!75!black,title=Instruction: Game of 24] 
Use numbers and basic arithmetic operations (+ - * /) to obtain 24.
\end{tcolorbox}

\begin{tcolorbox}[colback=gray!5!white,colframe=gray!75!black,title=Prompt: Game of 24] 
\textbf{Input:} 2 9 10 12

\textbf{Steps}:

12 * 2 = 24 (left: 9 10 24) Expression: 9, 10, (12) * (2)

10 - 9 = 1 (left: 24 1) Expression: (12) * (2), (10) - (9)

1 * 24 = 24 (left: 24) Expression: ((10) - (9)) * ((12) * (2)) 

Answer: (12 * 2) * (10 - 9) = 24
\end{tcolorbox}

\begin{tcolorbox}[colback=blue!5!white,colframe=blue!75!black,title=Revision: Game of 24] 
Using the given [input] numbers and basic arithmetic operations (+, -, *, /), follow the steps strictly to achieve a result of 24. 

All the [input] numbers can reach 24 by basic arithmetic operations (+, -, *, /).

If the final answer is not exactly 24, then the corresponding [Steps] is considered [wrong]. Please help me identify the exact wrong step based on its left number, among [Step 1, Step 2, Step 3]. If you are uncertain about which step is wrong, please begin your analysis with [Step 1] for better understanding.

\textbf{Input:} 2 9 10 12

\textbf{Steps:}

[Steps 1] 12 * 2 = 24 (left: 9 10 24) Expression: 9, 10, (12) * (2)

[Steps 2] 24 - 10 = 14 (left: 9 14) Expression: 9, ((12) * (2)) - (10)

[Steps 3] 9 + 14 = 23 (left: 23) Expression: (9) + ((12) * (2)) - (10)

The Steps are wrong. Because it can not reach 24 in the end. To be specific, 

23 is not equal to 24. [Steps 2] is wrong. Because it is impossible to reach 24 from the step 2. After Step 2, left numbers are 9, 14. 

9 + 14 = 23

9 * 14 = 126

9 -  14 = -5

It is impossible to reach 24 from [Steps 2].
\end{tcolorbox} 

\begin{tcolorbox}[breakable, colback=black!5!white,colframe=black!75!black,title=Instruction: 8-Puzzle] 
You are a virtual expert in solving a 8-puzzle problrm. Please follow the instructions and rules below to complete the solving. Your goal is to reach the goal state with valid moves.

[The goal state]

0 1 2 

3 4 5 

6 7 8

[Instructions]

The 8-puzzle consists of a 3x3 grid containing 8 numbered tiles (from 1 to 8) and one empty space (denoted by 0). 
Only 0 can be moved horizontally or vertically, and the objective is to reach the goal state from a given initial state. 
The goal state is typically the numbers ordered sequentially, with the 0 in the first position:

[The goal state]

0 1 2 

3 4 5 

6 7 8 
 
[Rules]

1. Only 0 can be moved horizontally or vertically.

2. Each move is chosen from the following set of options:

- 'Left': move 0 to the left

- 'Down': move 0 downward

- 'Right': move 0 to the right

- 'Up': move 0 upward

For example:

Before move:

1 2 3  

4 0 6  

7 8 5 

After move 'Left':

1 2 3  

0 4 6 

7 8 5 

Before move:

1 2 3  

4 0 6  

7 8 5 

After move 'Down':

1 2 3  

4 8 6  

7 0 5

Before move:

1 2 3  

4 0 6  

7 8 5 

After move 'Right':

1 2 3  

4 6 0  

7 8 5 

Before move:

1 2 3  

4 0 6  

7 8 5 

After move 'Up':

1 0 3  

4 2 6  

7 8 5

3. The next move must be chosen from the valid move set depending on the position of '0'.

For example:

p1  p2  p3 

p4  p5  p6 

p7  p8  p9 

(1) If '0' is located at position 'p1', the valid move set is ['Right', 'Down'].

(2) If '0' is located at position 'p2', the valid move set is ['Left', 'Right', 'Down'].

(3) If '0' is located at position 'p3', the valid move set is ['Left', 'Down'].

(4) If '0' is located at position 'p4', the valid move set is ['Right', 'Up', 'Down'].

(5) If '0' is located at position 'p5', the valid move set is ['Left', 'Right', 'Up', 'Down'].

(6) If '0' is located at position 'p6', the valid move set is ['Left', 'Up', 'Down'].

(7) If '0' is located at position 'p7', the valid move set is ['Right', 'Up'].

(8) If '0' is located at position 'p8', the valid move set is ['Left, 'Right', 'Up'].

(9) If '0' is located at position 'p9', the valid move set is ['Left', 'Up'].

4. Diagonal moves are not allowed.

5. The objective is to return the moves which can reach the goal state.

\end{tcolorbox}

\begin{tcolorbox}[breakable, colback=gray!5!white,colframe=gray!75!black,title=Prompt: 8-Puzzle] 
All given problems can be solved within 1 to 9 steps. The next move must be chosen from the valid move set. The maximum step number you can take is 9. Try to reach the goal state using the least number of steps ($\le$9). **DO NOT exceed 9 steps.**

\textbf{[Initial State]:}

3 1 2

6 4 5

7 8 0

\textbf{[Process]:}

3 1 2

6 4 5

7 8 0

\textbf{Step 1:} Choose one valid move from: [Left, Up]

Move: Left

Current State:

3 1 2

6 4 5

7 0 8

\textbf{Step 2:} Choose one valid move from: [Left, Right, Up]

Move: Left

Current State:

3 1 2

6 4 5

0 7 8

\textbf{Step 3:} Choose one valid move from: [Right, Up]

Move: Up

Current State:

3 1 2

0 4 5

6 7 8

\textbf{Step 4:} Choose one valid move from: [Right, Up]

Move: Up

Current State:

0 1 2

3 4 5

6 7 8

\textbf{Finished.}

\textbf{[Moves]:}

Left, Left, Up, Up
\end{tcolorbox}  

\begin{tcolorbox}[breakable, colback=blue!5!white,colframe=blue!75!black,title=Revision: 8-Puzzle] 
The given [Process] is not correct since it does not reach the goal state in the end. 

If the final answer does not reach the goal state, then the corresponding [Process] is considered [wrong]. Please help me identify the exact wrong step based on its left number, among [Step 1, Step 2, Step 3, ...]. If you are uncertain about which step is wrong, please begin your analysis with [Step 1] for better understanding.

Please help me identify the exact step number that is wrong. You must provide one wrong step.

\textbf{[Initial State]:}

3 1 2

6 4 5

7 8 0

\textbf{[Process]}

3 1 2

6 4 5

7 8 0

\textbf{Step 1:} Choose one valid move from: [Left, Up]

Left

3 1 2

6 4 5

7 0 8

\textbf{Step 2:} Choose one valid move from: [Left, Right, Up]

Left

3 1 2

6 4 5

0 7 8

\textbf{Step 3:} Choose one valid move from: [Right, Up]

Up

3 1 2

0 4 5

6 7 8

\textbf{Step 4:} Choose one valid move from: [Right, Up]

Right

3 1 2

4 0 5

6 7 8

Finished.

The given [Process] is not correct because number 3, 4, 0, 5 are not their goal positions in the end. The puzzle has failed on reaching its goal state.

Now please help me identify the exact step number that is wrong. You must provide one wrong step. If you can not provide an exact step number, please consider that it could be "all steps are wrong".

[Step 4] is wrong, with Move: Right.
\end{tcolorbox}  

\begin{tcolorbox}[breakable, colback=black!5!white,colframe=black!75!black,title=Instruction: Pocket Cube] 
You are a virtual expert in solving a 2x2 Pocket Cube. Your task is to restore a scrambled 2x2 Rubik's Cube to its original state. All the given problems can be solved in 1 to 4 moves. You cannot exceed more than 11 moves. Provide the sequence of moves required for the restoration. Please follow the instructions and rules below to complete the solving:

1. A 2x2 Pocket Cube has six faces, namely: [Upper, Front, Bottom, Left, Right, Back] Each consisting of a 2x2 grid of squares, with each square having its own color.

2. Colors in the Cube are represented in numbers: [0, 1, 2, 3, 4, 5]

3. The Cube's state is represented into a facelets expanding graph, for instance:

Upper: 

0 0 

0 0

Front: 

5 5 

2 2

Down: 

3 3 

3 3

Left: 

1 1 

4 4

Right: 

4 4 

1 1

Back: 

2 2 

5 5

4. A restoration of a Pocket Cube is to move squares in each face to have same numbers. Some example Restored States are:

[Restored State]

Upper: 

0 0 

0 0

Front: 

2 2 

2 2

Down: 

3 3 

3 3

Left: 

4 4 

4 4

Right: 

1 1 

1 1

Back: 

5 5 

5 5

Or

[Restored State]

Upper: 

2 2 

2 2

Front: 

0 0

0 0

Down: 

3 3 

3 3

Left: 

1 1 

1 1

Right: 

4 4 

4 4

Back: 

5 5 

5 5

You must make move to the Cube to achieve a Restored State, not limited to the above one. Note that we just need each face to have same numbers, no matter which face has which color.

5. You are only allowed to use following moves [U, U', U2, R, R', R2, F, F', F2]. 

["U": Turn the Upper face of the cube 90 degrees clockwise.
For instance, after taking move U:

Upper: 

0 0 

0 0

Front: 

2 2 

2 2

Down: 

3 3 

3 3

Left: 

4 4 

4 4

Right: 

1 1 

1 1

Back: 

5 5 

5 5

will become

Up:

0 0

0 0

Front:

1 1

2 2

Down:

3 3

3 3

Left:

2 2
4 4

Right:

5 5

1 1

Back:

4 4
5 5

"U'": Turn the Upper face of the cube 90 degrees counterclockwise (or anti-clockwise). For instance, after taking move U':

Upper: 

0 0 

0 0

Front: 

2 2 

2 2

Down: 

3 3 

3 3

Left: 

4 4 

4 4

Right: 

1 1 

1 1

Back: 

5 5 

5 5

will become

Upper:

0 0

0 0

Front:

4 4

2 2

Down:

3 3

3 3

Left:

5 5

4 4

Right:

2 2

1 1

Back:

1 1

5 5

"U2": Turn the Upper face of the cube 180 degrees (a half turn). For instance, after taking move U2:

Upper: 

0 0 

0 0

Front: 

2 2 

2 2

Down: 

3 3 

3 3

Left: 

4 4 

4 4

Right: 

1 1 

1 1

Back: 

5 5 

5 5

will become

Up:

0 0

0 0

Front:

5 5

2 2

Down:

3 3

3 3

Left:

1 1

4 4

Right:

4 4

1 1

Back:

2 2

5 5
      
"R": Turn the Right face of the cube 90 degrees clockwise.
For instance, after taking move R:

Upper: 

0 0 

0 0

Front: 

2 2 

2 2

Down: 

3 3

3 3

Left: 

4 4 

4 4

Right: 

1 1 

1 1

Back: 

5 5 

5 5

will become

Upper:

0 2

0 2

Front:

2 3

2 3

Down:

3 5

3 5

Left:

4 4

4 4

Right:

1 1

1 1

Back:

0 5

0 5
      
"R'": Turn the Right face of the cube 90 degrees counterclockwise. For instance, after taking move R':

Upper: 

0 0 

0 0

Front: 

2 2 

2 2

Down: 

3 3 

3 3

Left: 

4 4 

4 4

Right: 

1 1 

1 1

Back: 

5 5 

5 5

will become

Upper:

0 5

0 5

Front:

2 0

2 0

Down:

3 2

3 2

Left:

4 4

4 4

Right:

1 1

1 1

Back:

3 5

3 5
      
"R2": Turn the Right face of the cube 180 degrees.
For instance, after taking move R':

Upper: 

0 0 

0 0

Front: 

2 2 

2 2

Down: 

3 3 

3 3

Left: 

4 4 

4 4

Right: 

1 1 

1 1

Back: 

5 5 

5 5

will become

Up:

0 3

0 3

Front:

2 5

2 5

Down:

3 0

3 0

Left:

4 4

4 4

Right:

1 1

1 1

Back:

2 5

2 5

"F": Turn the Front face of the cube 90 degrees clockwise.
For instance, after taking move F:

Upper: 

0 0 

0 0

Front: 

2 2 

2 2

Down: 

3 3 

3 3

Left: 

4 4 

4 4

Right: 

1 1 

1 1

Back: 

5 5 

5 5

will become

Up:

0 0

4 4

Front:

2 2

2 2

Down:

1 1

3 3

Left:

4 3

4 3

Right:

0 1

0 1

Back:

5 5

5 5

"F'": Turn the Front face of the cube 90 degrees counterclockwise.
For instance, after taking move F':
Upper: 

0 0 

0 0

Front: 

2 2 

2 2

Down: 

3 3 

3 3

Left: 

4 4 

4 4

Right: 

1 1 

1 1

Back: 

5 5 

5 5

will become

Upper: 

0 0 

1 1

Front: 

2 2 

2 2

Down: 

4 4 

3 3

Left: 

4 0 

4 0

Right: 

3 1 

3 1

Back: 

5 5 

5 5      
 
"F2": Turn the Front face of the cube 180 degrees.
For instance, after taking move F2:

Upper: 

0 0 

0 0

Front:

2 2 

2 2

Down: 

3 3 

3 3

Left: 

4 4 

4 4

Right: 

1 1 

1 1

Back: 

5 5 

5 5

will become

Upper:

0 0

3 3

Front:

2 2

2 2

Down:

0 0

3 3

Left:

4 1

4 1

Right:

4 1

4 1

Back:

5 5

5 5

\end{tcolorbox}

\begin{tcolorbox}[breakable, colback=gray!5!white,colframe=gray!75!black,title=Prompt: Pocket Cube] 
All the given problems can be solved in 1 to 4 moves. **You cannot exceed more than 11 moves.** Please complete [Process] and return the [Restoration Moves].

\textbf{[Initial Cube State]:}

Upper:

4 5

4 4

Front:

5 1

5 0

Down:

0 0

2 0

Left:

1 1

3 2

Right:

2 2

4 3

Back:

3 3

1 5

\textbf{[Process]:}

\textbf{[Step 1]}

[Move] R

[Current Cube State]

Upper:

4 0

4 0

Front:

5 5

0 1

Down:

0 1

2 2

Left:

1 1

3 3

Right:

2 2

4 3

Back:

4 3

5 5

\textbf{[Step 2]}

[Move] U'

[Current Cube State]

Upper:

0 0

4 4

Front:

0 1

0 1

Down:

2 2

2 2

Left:

1 1

3 3

Right:

4 3

4 3

Back:

5 5

5 5

\textbf{[Step 3]}

[Move] F'

[Current Cube State]

Upper:

0 0

0 0

Front:

1 1

1 1

Down:

2 2

2 2

Left:

3 3

3 3

Right:

4 4

4 4

Back:

5 5

5 5

\textbf{Finished.}

Now strictly follow the above process to form Restoration Moves.

\textbf{[Restoration Moves]:}

R U' F'
\end{tcolorbox}  

\begin{tcolorbox}[breakable, colback=blue!5!white,colframe=blue!75!black,title=Revision: Pocket Cube] 
The given [Process] is not correct since it does not reach the goal state in the end. 

If the final answer does not reach the goal state, then the corresponding [Process] is considered [wrong]. Please help me identify the exact wrong step based on its left number, among [Step 1, Step 2, Step 3, ...]. If you are uncertain about which step is wrong, please begin your analysis with [Step 1] for better understanding.

Please help me identify the exact step number that is wrong. You must provide one wrong step.

\textbf{[Initial Cube State]:}

Upper:

4 5

4 4

Front:

5 1

5 0

Down:

0 0

2 0

Left:

1 1

3 2

Right:

2 2

4 3

Back:

3 3

1 5

\textbf{[Process]:}

\textbf{[Step 1]}

[Move] R

[Current Cube State]

Upper:

4 0

4 0

Front:

5 5

0 1

Down:

0 1

2 2

Left:

1 1

3 3

Right:

2 2

4 3

Back:

4 3

5 5

\textbf{[Step 2]}

[Move] U'

[Current Cube State]

Upper:

0 0

4 4

Front:

0 1

0 1

Down:

2 2

2 2

Left:

1 1

3 3

Right:

4 3

4 3

Back:

5 5

5 5

\textbf{[Step 3]}

[Move] F2

[Current Cube State]

Upper:

0 0

1 1

Front:

2 2

2 2

Down:

4 4

3 3

Left:

4 0

4 0

Right:

3 1

3 1

Back:

5 5

5 5

Finished.

After finishing all the moves: The Upper face still has 2 differnet colors. The Down face still has 2 differnet colors. The Left face still has 2 differnet colors. The Right face still has 2 differnet colors. 

The given [Process] is not correct because not every face has the same numbers in the end. The cube has failed on restoring to its original state.
Now please help me identify the exact step number that is wrong. You must provide one wrong step. If you can not provide an exact step number, please consider that it could be "all steps are wrong".

[Step 3] is wrong, with Move: F2.
\end{tcolorbox}

\end{document}